%% file: main.tex
\documentclass[10pt,twocolumn,letterpaper]{article}

\usepackage{cvpr}              

\input{preamble}

\definecolor{cvprblue}{rgb}{0.21,0.49,0.74}
\usepackage[pagebackref,breaklinks,colorlinks,citecolor=cvprblue]{hyperref}
\usepackage{multirow}
\usepackage{tikz}
\usepackage[accsupp]{axessibility}

\newcommand{\ours}{PhysPT}

\def\paperID{12901} 
\def\confName{CVPR}
\def\confYear{2024}

\title{PhysPT: Physics-aware Pretrained Transformer for Estimating Human Dynamics from Monocular Videos}

\author{Yufei Zhang$^1$, Jeffrey O. Kephart$^2$, Zijun Cui$^{1,3}$, Qiang Ji$^1$ \\
$^1$Rensselaer Polytechnic Institute, $^2$IBM Research, $^3$University of Southern California \\
{\tt\small \{zhangy76, jiq\}@rpi.edu, kephart@us.ibm.com, ceejkl@gmail.com}
}
\begin{document}
\maketitle
\input{sec/0_abstract}

\input{sec/1_intro}
\input{sec/2_relatedwork}

\input{sec/3_method}

\input{sec/4_experiment}
\input{sec/5_conclusion}

{
    \small
    \bibliographystyle{ieeenat_fullname}
    \bibliography{main}
}

\input{sec/X_suppl}

\end{document}

%% file: preamble.tex
\usepackage[dvipsnames]{xcolor}


%% file: sec/0_abstract.tex
\begin{abstract}

While current methods have shown promising progress on estimating 3D human motion from monocular videos, their motion estimates are often physically unrealistic because they mainly consider kinematics. In this paper, we introduce \textbf{Phys}ics-aware \textbf{P}retrained \textbf{T}ransformer (\textbf{\ours}), which improves kinematics-based motion estimates and infers motion forces. {\ours} exploits a Transformer encoder-decoder backbone to effectively learn human dynamics in a self-supervised manner. Moreover, it incorporates physics principles governing human motion. Specifically, we build a physics-based body representation and contact force model. We leverage them to impose novel physics-inspired training losses (i.e., force loss, contact loss, and Euler-Lagrange loss), enabling {\ours} to capture physical properties of the human body and the forces it experiences. Experiments demonstrate that, once trained, {\ours} can be directly applied to kinematics-based estimates to significantly enhance their physical plausibility and generate favourable motion forces. Furthermore, we show that these physically meaningful quantities translate into improved accuracy of an important downstream task: human action recognition.

\end{abstract}

%% file: sec/1_intro.tex
\section{Introduction}
\label{sec:intro}

Monocular 3D human motion estimation is essential for applications like human-computer interaction~\cite{teo2019mixed,chakraborty2018review}, motion analysis~\cite{colombel2020physically}, and robotics~\cite{erickson2020assistive}. This task is inherently challenging due to the absence of depth information and the intricate interplay of forces and human body movements. 

Recent advances in deep learning, along with progress in 3D human modeling~\cite{loper2015smpl, xu2020ghum}, have substantially improved the reconstruction of 3D humans from a single image~\cite{kanazawa2018end, kolotouros2019learning, li2021hybrik, tripathi20233d, zhang2023body}. With video inputs, current research aims to enhance model performance by exploiting temporal information. Some authors devise
temporal models that extract meaningful features from videos to improve performance~\cite{hossain2018exploiting, pavllo20193d, cai2019exploiting, wang2020motion, cheng2020graph, cheng2021monocular, zeng2021learning, lee2021uncertainty, choi2021beyond, tang20233d, shen2023global, zhang2023two,shin2023wham}. Other authors learn motion priors that capture natural 3D body movement patterns. Integrating the learned priors into model training can promote smooth motion estimates~\cite{kocabas2020vibe,luo20203d,rempe2021humor,shi2023phasemp}. While these approaches enhance reconstruction to some extent, they often produce unrealistic estimates characterized by noticeable physical artifacts such as motion jittering and foot skating.

To address this limitation, a promising strategy is to leverage physical principles governing body movements. In this approach, the human body is treated as an articulated rigid body, and human dynamics are described through the Euler-Lagrange equations. These equations link body mass, inertia, and physical forces (including joint actuations and contact forces) to body motion through ordinary differential equations. Some researchers~\cite{li2019estimating,shimada2020physcap,rempe2020contact, xie2021physics,gartner2022trajectory,gartner2022differentiable,yang2023ppr} formulate optimization frameworks that jointly estimate unknown physical parameters and refine kinematics-based estimates by aligning them with the physics equations. Alternatively, others~\cite{yuan2021simpoe,huang2022neural,luo2022embodied,shimada2021neural,li2022d} employ learning-based frameworks. They sidestep the cumbersome manual parameter tuning inherent in optimization-based methods by training neural networks to predict the parameters.

Yet, a key challenge remains: physics information, including physical properties of human bodies and motion forces, is absent in current 3D motion capture datasets~\cite{mahmood2019amass}. To incorporate physics, existing methods generally rely on physics engines~\cite{todorov2012mujoco,coumans2016pybullet}. This entails creating proxy bodies with geometric primitives to capture body properties, importing these proxies into a physics engine, and then leveraging the physics engine to compute the necessary physical parameters and simulate body motion. The problem with this approach arises from the difficulty of efficiently computing gradients from physics engine outputs~\cite{gartner2022differentiable}, thereby limiting their seamless integration with deep learning models. Moreover, existing physics-based models are primarily trained with 3D annotated videos, which are challenging to acquire in practice. Consequently, the trained models may not generalize well to unseen scenarios. 

In this paper, we propose a novel framework for learning human dynamics that circumvents the need for 3D annotated videos and effectively integrates physics with advanced deep models. Specifically, drawing inspiration from recent success of pre-trained Transformers~\cite{vaswani2017attention} in temporal modeling, we propose leveraging a Transformer encoder-decoder architecture and training in a self-supervised manner by reconstructing input human motion. When incorporating physics, we bypass unrealistic body proxies by directly computing body physical properties from the widely adopted 3D body model, SMPL~\cite{loper2015smpl}. We also introduce a contact model to effectively model the contact forces. We utilize these models to derive motion forces from training sequences and impose novel physics-inspired training losses, including force loss, contact loss, and Euler-Lagrange loss. We train the Transformer model only using existing motion capture data. Once trained, our physics-aware pretrained Transformer ({\ours}) can be applied on top of any kinematics-based reconstruction model to produce enhanced motion and force estimates from monocular videos. In summary, our main contributions include:
\begin{itemize}
    \item We introduce {\ours}, a Transformer encoder-decoder model trained through self-supervised learning with incorporation of physics. Once trained, {\ours} can be combined with any kinematic-based model to estimate human dynamics without additional model fine-tuning.
    \item We present a novel framework for incorporating physics. This includes a physics-based body representation and a contact force model, and, subsequently, the imposition of a set of novel physics-inspired losses for model training.  
    \item We demonstrate through experiments that PhysPT significantly enhances the physical plausibility of motion estimates and infers favourable motion forces. Furthermore, we demonstrate that the enhanced motion and force estimates translate into accuracy improvements in an important downstream task: human action recognition. 
\end{itemize}

%% file: sec/2_relatedwork.tex
\section{Related Work}
\label{sec:relatedwork}

\noindent\textbf{Kinematics-based Human Motion Estimation.} Methods modeling body kinematics estimate body geometry configuration solely. Among these approaches, one line of work involves optimization-based pipelines that iteratively fit a prior body model to 2D observations to reconstruct a 3D human~\cite{zhou2016sparseness,bogo2016keep,arnab2019exploiting,xiang2019monocular,vo2020spatiotemporal,guan2021bilevel,ye2023decoupling}. Others embrace deep learning models to directly predict 3D human bodies. Given a single input image, existing methods have proposed different model
architectures with various intermediate and output representations to improve the reconstruction accuracy~\cite{moon2020i2l,li2021hybrik,kolotouros2021probabilistic,kocabas2021pare,lin2021end,li2022cliff,yoshiyasu2023deformable,sengupta2023humaniflow,ma20233d,fang2023learning,zhang20233d,wang2023zolly}. 

Given input of a monocular video, current kinematics-based methods aim to fully harness temporal information to obtain improved results. Various temporal models are developed based on Temporal Convolutional Networks~\cite{pavllo20193d,kanazawa2019learning,lee2021uncertainty,zeng2022smoothnet,zhang2023two}, Graph Convolutional Networks~\cite{cai2019exploiting,wang2020motion,cheng2020graph,cheng2021monocular,zeng2021learning}, Recurrent Neural Networks~\cite{hossain2018exploiting,kocabas2020vibe,luo20203d,choi2021beyond,you2023co,sun2023trace}, Transformer~\cite{rajasegaran2021tracking,li2022deep,zhu2022motionbert,zhao2023poseformerv2,tang20233d,shen2023global,qiu2023psvt,goel2023humans}, or those explicitly capturing and exploiting attention~\cite{sun2019human,liu2020attention,wan2021encoder,wei2022capturing}.
Another common approach to encouraging realistic temporal predictions is to incorporate smoothness constraints or motion priors during training~\cite{kocabas2020vibe,luo20203d,rempe2021humor,zhang2021learning,yuan2022glamr}. These kinematics-based methods, however, often produce noticeable physical artifacts due to their failure to realistically capture the complexity of human motion.

\noindent\textbf{Physics-based Human Motion Estimation.} Physics-based approaches explicitly leverage physics principles, particularly the Euler-Lagrange equations, to capture human dynamics. Prior works have adopted optimization frameworks to jointly estimate motion forces and refine initial kinematics-based motion estimates by minimizing the residuals introduced by the Euler-Lagrange equations~\cite{li2019estimating,rempe2020contact,xie2021physics}. Directly estimating the exerted forces is challenging; therefore, others employ a character control methodology. In this paradigm, kinematics-based estimates act as reference motions, and the forces needed to emulate these motions in a physics engine are predicted by estimating the parameters of a controller~\cite{zheng2015human,levine2012physically,shimada2020physcap,yi2022physical,gartner2022differentiable,yang2023ppr}. However, these optimization-based methods often require careful tuning of the control parameters. Some approaches instead leverage neural networks to estimate the parameters, where the models are trained through fully supervised learning~\cite{shimada2021neural} or reinforcement learning~\cite{yuan2021simpoe,huang2022neural,luo2022embodied,ju2023physics}. In these approaches, incorporating the physics engine alongside learning models falls short of achieving an effective end-to-end integration of physics. Li \textit{et al.}~\cite{li2022d} enhance the learning process by analytically computing some of the physical parameters coupled with the usage of 3D supervisions. While the produced results are promising, existing learning-based methods rely on 3D annotated videos for training and often exhibit poor generalization. In contrast, our model adopts self-supervised learning, trained solely using existing 3D motion data without images. Additionally, we introduce a novel framework that seamlessly bridges the gap between body kinematics and physics without relying on physics engines, facilitating the effective integration of physics with advanced deep learning models.

\noindent\textbf{Full Human Dynamics Estimation.} Fully capturing human dynamics requires determining both body movements and the forces exerted by individuals~\cite{sherman2011simbody,bidermaninverse,scott2020image,chiquier2023muscles}. Prevailing methods for estimating human dynamics primarily focus on inferring forces from 3D motion capture data~\cite{brubaker2009estimating,zell2017learning,zell2020learning, zell2020weakly}. These estimated forces are used to facilitate tasks such as human action recognition~\cite{mansur2012inverse} or human motion prediction~\cite{liu2022dynamic,zhang2022pimnet,zhang2024incorporating}. Our approach can address a more challenging task: the estimation of full human dynamics from a monocular video. We do so without utilizing any ground truth force labels. To our knowledge, we are the first to demonstrate that forces inferred from monocular videos can improve human action recognition.

%% file: sec/3_method.tex
\begin{figure*}[t]
    \centering
    \includegraphics[width=0.99\linewidth]{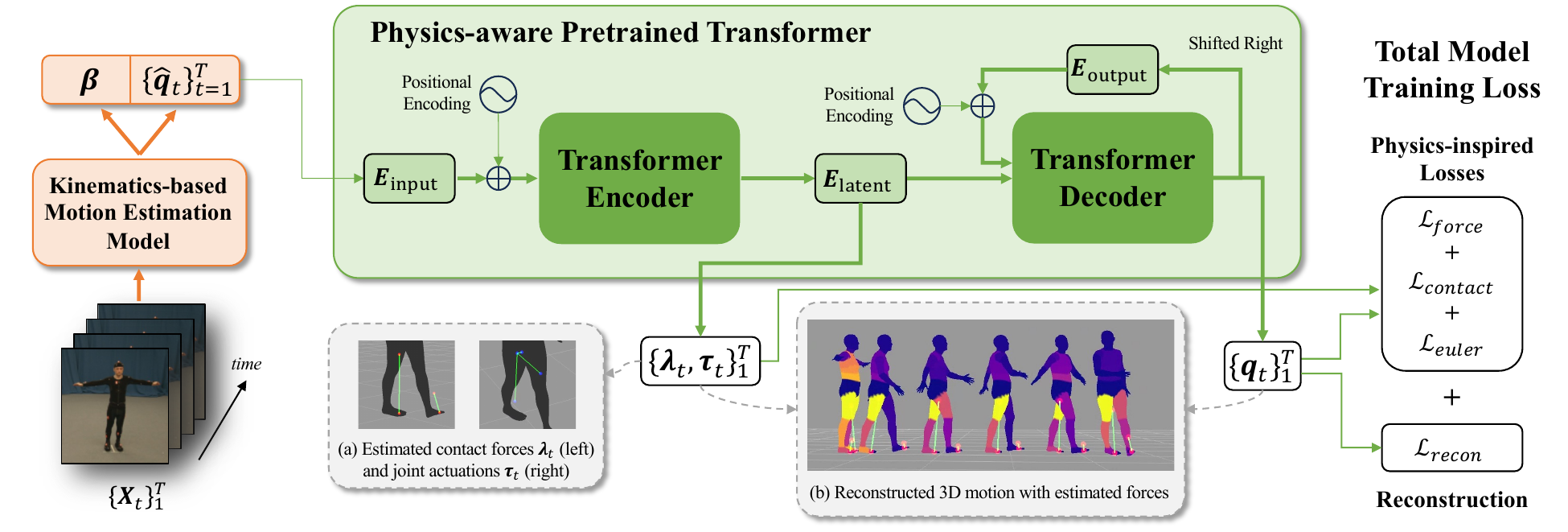}
    \vspace{-0.25cm}
    \caption{\textbf{Method Overview.} The proposed framework consists of a kinematics-based motion estimation model (orange) and a physics-aware pre-trained Transformer (green) for estimating human dynamics from a monocular video. Inset (a) illustrates joint actuation of right pelvis and contact forces at each foot. (b) illustrates reconstructed body motion and inferred forces with lighter colors representing greater joint actuation magnitudes (e.g. upper body joints when the figure is standing, and leg joints when it is walking).}
    \label{fig:overview}
\end{figure*}

\section{Proposed Method}

Fig.~\ref{fig:overview} shows an overview of our method. Given $T$ video frames $\{\mathbf{X}_t\}_{t=1}^T$, a kinematics-based model is employed to generate initial motion estimates $\{\hat{\mathbf{q}}_t\}_{t=1}^{T}$, followed by the proposed {\ours} to estimate refined motion $\{\mathbf{q}_t\}_{t=1}^{T}$ and infer forces $\{\boldsymbol{\lambda}_t,\boldsymbol{\tau}_t\}_{t=1}^{T}$. In Sec.~\ref{sec:preliminaries}, we introduce the physics equations for modeling human dynamics and describe how we generate the kinematics-based estimates. In Sec.~\ref{sec:physicsbranch}, we delve into the key components of {\ours}.

\subsection{Preliminary}

\label{sec:preliminaries}

\noindent\textbf{Euler-Lagrange Equations.} As a complex physical system composed of multiple interacting body parts, the human body is often modeled using rigid body dynamics. In this context, the Euler-Lagrange equations provide a concise mathematical description of human dynamics within a generalized coordinate system. 

The generalized coordinates are defined by variables that fully specify the system's state. Based on the successful geometry body model SMPL~\cite{loper2015smpl}, we represent a 3D human body in terms of a mesh model using body pose $\boldsymbol{\theta}\in\mathbb{R}^{23\times3}$ and body shape parameters $\boldsymbol{\beta}\in\mathbb{R}^{10}$. The pose parameters characterize the rotations of 23 body joints, while the shape parameters control the variations in body attributes, e.g.\ girth. Given that the body shape remains constant within a video, the 3D body trajectory in the world frame can be fully specified by the pose $\boldsymbol{\theta}$ plus body translation $\mathbf{T}\in\mathbb{R}^{3}$ and rotation $\mathbf{R}\in\mathbb{R}^{3}$ via a generalized coordinate $\mathbf{q}$: 
\begin{equation}
\label{eq:generalizedposition}
    \mathbf{q}=\{\mathbf{T},\mathbf{R},\boldsymbol{\theta}\}.
\end{equation}

Denoting the velocity and the acceleration in the generalized coordinates as $\dot{\mathbf{q}}$ and  $\ddot{\mathbf{q}}$, respectively, the body dynamics governed by the Euler-Lagrange equations are:
\begin{equation}
\label{eq:EulerLag}
\resizebox{.9\hsize}{!}{$
    \mathbf{M}\big(\mathbf{q} ;\ \mathbf{m},\mathbf{I}\big)\ddot{\mathbf{q}} +\mathbf{C}\big(\mathbf{q},\dot{\mathbf{q}};\ \mathbf{m},\mathbf{I}\big) + \mathbf{g}\big(\mathbf{q};\ \mathbf{m}\big)= \mathbf{J}_C^T\boldsymbol{\lambda} + \boldsymbol{\tau},
    $}
\end{equation}
where $\mathbf{M}\big(\mathbf{q} ;\ \mathbf{m},\mathbf{I}\big)$ is the generalized inertia matrix determined by the position $\mathbf{q}$, the body mass $\mathbf{m}$, and the inertia $\mathbf{I}$. $\mathbf{C}\big(\mathbf{q},\dot{\mathbf{q}};\ \mathbf{m},\mathbf{I}\big)$ represents the Coriolis and centrifugal forces. $\mathbf{g}\big(\mathbf{q};\ \mathbf{m}\big)$ indicates the gravitational forces. $\boldsymbol{\lambda}\in\mathbb{R}^{3n_c}$ denotes the contact forces, where $n_c$ is the number of points of contact. $\mathbf{J}_C\in\mathbb{R}^{3n_c\times 75}$ is the contact Jacobian matrix that describes the mapping between the contact points' Cartesian velocity, $\mathbf{v}_C\in\mathbb{R}^{3n_c}$, and the generalized velocity, $\dot{\mathbf{q}}$, according to the equation  $\mathbf{v}_C=\mathbf{J}_C\dot{\mathbf{q}}$. Additionally, $\boldsymbol{\tau}\in\mathbb{R}^{75}$ represents joint actuations, as exemplified in Fig.~\ref{fig:overview}-a for right pelvis joint.

\noindent\textbf{Kinematics-based Motion Estimation Model.} We first employ an established method to obtain per-frame 3D body pose and shape $\{{\hat{\boldsymbol{\theta}}_t,\hat{\boldsymbol{\beta}}_t}\}_{t=1}^{T}$ from the video input. It places no restrictions on which method is used; for our experiments we use recent publicly-available models. The pose and shape estimated by those traditional 3D human reconstruction models only capture body movements in the body frame, lacking a global motion trajectory to fully specify the generalized positions defined in Eq.~\ref{eq:generalizedposition}. As in \cite{yuan2022glamr}, we train a global trajectory predictor to provide per-frame global translation and rotation $\{{\hat{\mathbf{T}}_t,\hat{\mathbf{R}}_t}\}_{t=1}^{T}$ based on the local body movements. The global trajectory predictor is trained independently and produces the global estimates without additional model fine-tuning. Further details of the model architecture and training are in Supp.~\ref{supp:globaltraj}. Finally, by combining the estimated global trajectory with the local body pose, we obtain the initial generalized position estimates $\{\hat{\mathbf{q}}_t\}_{t=1}^{T}$, which are input to {\ours} for further refinement. Meanwhile, we consider the final shape estimate $\boldsymbol{\beta}=\frac{1}{T}\sum_{t=1}^T\hat{\boldsymbol{\beta}}_t$ since the subject's shape remains unchanged over time.

\subsection{Physics-aware Pretrained Transformer}
\label{sec:physicsbranch}

The initial kinematics-based motion estimates maintain reasonable per-frame reconstruction accuracy. The Physics-aware Pretrained Transformer introduced in this section further enhances the motion estimates and infers motion forces. In the following, we first introduce the Transformer encoder-decoder backbone of {\ours} in Sec.~\ref{sec:transformer}. To incorporate physics into the model, we build a physics-based body representation (Sec.~\ref{sec:bodyrepresentation}) and a contact force model (Sec.~\ref{sec:forcelabel}), which enable the formulation of physics-inspired training losses (Sec.~\ref{sec:physicsincorp}).

\subsubsection{Transformer Encoder-Decoder Backbone}
\label{sec:transformer}

Differing from existing works that primarily utilize a Transformer encoder to learn representations, we exploit a Transformer encoder-decoder architecture~\cite{vaswani2017attention}. As illustrated in Fig.~\ref{fig:overview} (green region), the model first extracts embedding $\mathbf{E}_{input}\in \mathbf{R}^{T\times n_f}$ from the kinematics-based estimates $\{\hat{\mathbf{q}}_t\}_{t=1}^{T}$ using a linear layer. This $\mathbf{E}_{input}$, combined with a time positional encoding, is then fed into the Transformer encoder to generate a latent embedding $\mathbf{E}_{latent}\in\mathbf{R}^{T\times n_l}$. Here, $n_f$ and $n_l$ are embedding dimensions. Subsequently, the decoder generates refined estimates $\{\mathbf{q}_t\}_{t=1}^{T}$ via autoregressive prediction. Specifically, at time frame $m+1$, the previous $m$ estimates are projected into embeddings $\mathbf{E}_{output}\in\mathbf{R}^{m\times n_f}$, to which a positional encoding is added. Together with $\mathbf{E}_{latent}$, this is input to the Transformer decoder to produce the motion prediction.

Leveraging the Transformer encoder-decoder backbone can effectively capture temporal information in a self-supervised manner by reconstructing the input. Specifically, denoting an input sequence from existing 3D motion capture data as $\{\bar{\mathbf{q}}_t\}_{t=1}^{T}$, we compute a mean squared error on the generalized positions and 3D joint positions, leading to the added reconstruction loss $\mathcal{L}_{recon}$: 
\begin{equation}
\label{eq:recon}
\resizebox{.54\hsize}{!}{$
\begin{split}
    \mathcal{L}_{recon} &= \sum_{t=1}^T\gamma_{q}\mathcal{L}_{q,t} + \gamma_{J}\mathcal{L}_{J,t}, \\
    \mathcal{L}_{q,t}&=\|\mathbf{q}_t-\bar{\mathbf{q}}_t\|_2^2, \\
    \mathcal{L}_{J,t}&=\|\mathbf{J}_t-\bar{\mathbf{J}}_t\|_2^2, \\
\end{split}$}
\end{equation}
where the 3D joint positions $\mathbf{J}_t\in\mathbb{R}^{n_J\times3}$ are computed from the generalized positions and the body shape parameters using forward kinematics and $n_J$ is the number of body joints. $\gamma_{q}$ and $\gamma_{J}$ are training loss weights. To enhance model robustness, we introduce random Gaussian noise into the input during training while the model is still tasked with reconstructing the clean input. Up to this point, the Transformer model is trained to effectively learn the geometry information from motion data, but it is agnostic to physics and insufficient to faithfully capture human dynamics.

\subsubsection{Physics-based Body Representation}
\label{sec:bodyrepresentation}

\begin{figure}[t]
    \centering
    \includegraphics[width=0.46\textwidth]{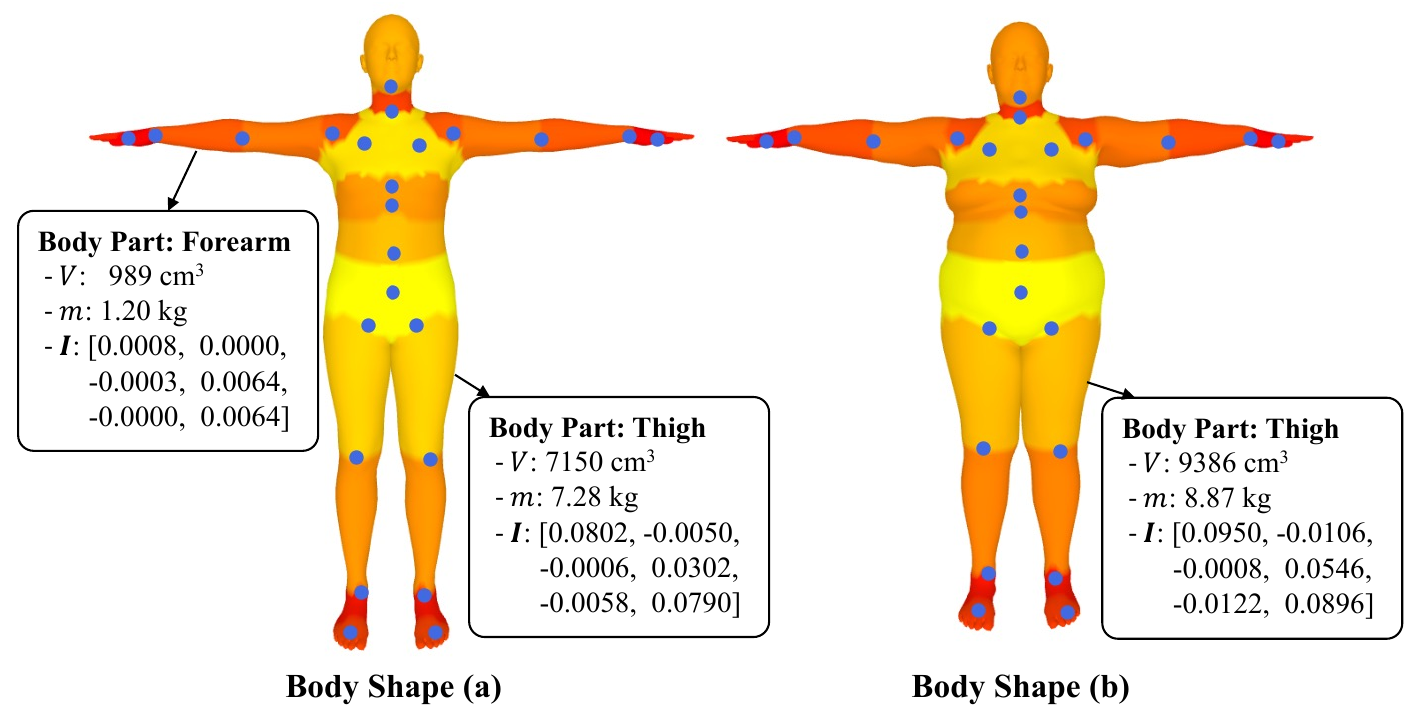}
    \vspace{-0.25cm}
    \caption{\textbf{Phys-SMPL.} Besides 3D positions, Phys-SMPL models the volume ($V$), mass ($m$), and inertia ($\mathbf{I}$) of every body parts. Lighter colors represent larger body weight distributions.}
    \label{fig:bodyrepresentation}
\end{figure}

To empower the model to capture physics, we need to first model the physical properties of human bodies. For this purpose, we introduce a physics-based body representation, Phys-SMPL. As shown in Fig.~\ref{fig:bodyrepresentation}, Phys-SMPL incorporates the mass $\mathbf{m}(\boldsymbol{\beta})\in \mathbb{R}^{24}$ and inertia $\mathbf{I}(\boldsymbol{\beta})\in \mathbb{R}^{24\times3\times3}$ of 24 body parts in additional to SMPL's geometry information. Specifically, we first close the meshes of each body part. This allows us to compute the volume of a body part as the sum of the tetrahedrons formed by its centroid and mesh faces. Based on the average mass density of the human body~\cite{plagenhoef1983anatomical}, we calculate the mass and, subsequently, the inertia of different body parts. Note that these physical body properties are computed directly from SMPL's geometry information specified by the shape parameters $\boldsymbol{\beta}$, without the need for creating unrealistic body proxies. Expanding on Phys-SMPL, we analytically calculate the physical terms in the Euler-Lagrange equations (Eq.~\ref{eq:EulerLag}). The analytical computation of physical parameters is fully differentiable, enabling the seamless integration of physics with learning models during the training process. Further details of Phys-SMPL and the analytical calculation are in Supp.~\ref{supp:physicsmodel}.

\subsubsection{Continuous Contact Force Model} 
\label{sec:forcelabel}

\begin{figure}[t]
    \centering
    \includegraphics[width=0.3\textwidth]{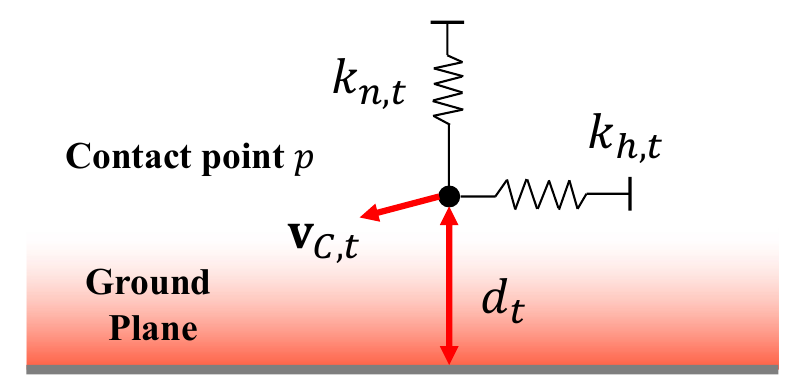}
    \vspace{-0.25cm}
    \caption{\textbf{Continuous Contact Force Model.} The contact forces received by a point $p$ at time frame $t$ are determined by its velocity and distance to the ground through a spring-mass system built along the horizontal ($k_{h,t}$) and normal ($k_{n,t}$) directions.}
    \label{fig:continuouscontact}
\end{figure}

To capture human dynamics, the motion forces must be modeled as well. For the joint actuations and contact forces, modeling the contact forces can be particularly challenging. The contact status often needs to be determined beforehand --- and this is in itself difficult to do accurately. The discrete contact status also introduces a non-differentiable process in estimating the forces. To address this issue, we draw inspiration from the continuous contact model proposed by \cite{brubaker2009estimating} for estimating ground reaction forces from 3D motion, which entails introducing a spring-mass system as illustrated in Fig.~\ref{fig:continuouscontact}. Specifically, the ground contact force experienced by a point $p$ at time $t$ is modeled as:
\begin{equation}
\label{Eq:springmass}
\boldsymbol{\lambda}_{p,t} = s_{p,t} (-k_{h,t}\mathbf{b}_{h,t}-
    k_{n,t}\mathbf{b}_{n,t}-c_t\mathbf{v}_{C,t}).
\end{equation}
where $k_{h,t}$ and $k_{n,t}$ denote the stiffness of the spring-mass in the horizontal and normal directions, respectively, while $c_t$ represents the damping factor. The scalar $s_{p,t}=2\sigma(-60d_t)\sigma(-60\|\mathbf{v}_{C,t}\|)$ regulates the force magnitude, where $\sigma(\cdot)$ represents the Sigmoid function, $d_t$ denotes the point's distance to the ground, and $\|\mathbf{v}_{C,t}\|$ is its velocity. Additionally, $\mathbf{b}_{h,t}=[d_{t,x}-0.5;d_{t,y}-0.5;0]$ and $\mathbf{b}_{n,t}=[0;0;d_t-2]$ is the distance to the reference point in the horizontal and normal direction, respectively. Here, $d_{t,x}$ and $d_{t,y}$ are projections of $d_t$ onto the $x$ and $y$ axes using the normal of the contact point. The units are in meters. For the sake of computational efficiency (and as further discussed in Supp.~\ref{supp:contactregion}), we apply the contact model to a subset of vertices within each body part. The contact model captures the essentials of natural contact behavior, where points closest to the ground and most stable experience larger forces. Utilizing the contact model also avoids the problems presented in estimating the discrete contact status.

\subsubsection{Physics-inspired Training Losses}
\label{sec:physicsincorp}

Building upon the physics-based body representation and force model, we can effectively integrate physics with the model by utilizing several physics-inspired training losses.

To formulate these losses, the first step involves deriving valuable motion force information from training sequences. Given a 3D trajectory $\{\bar{\mathbf{q}}_t\}_{t=1}^T$ from training data, we utilize the finite difference to obtain the velocity and acceleration $\{\dot{\bar{\mathbf{q}}}_t,\ddot{\bar{\mathbf{q}}}_t\}_{t=1}^{T}$. We then formulate the following optimization problem to recover the motion forces at time frame $t$ as: 
\begin{equation}
\label{Eq:pseudoforce}
\resizebox{.98\hsize}{!}{$
\begin{split}
    \arg\min_{\mathbf{x}_t,\boldsymbol{\tau}_t} \quad &  \|  \bar{\mathbf{M}}_t\ddot{\bar{\mathbf{q}}}_t +\bar{\mathbf{C}}_t + \bar{\mathbf{g}}_t - \bar{\mathbf{J}}_{C,t}^T\bar{\mathbf{A}}_t\mathbf{x}_t - \boldsymbol{\tau}_t \|_2^2 \\
    s.t. \quad & \mathbf{0}<\mathbf{x}_t<\bar{\mathbf{x}}_{max}. \quad  \text{(stiffness and damping constraints)} 
\end{split}$}
\end{equation}
The objective is a least squared error introduced by the Euler-Lagrange equations in Eq.~\ref{eq:EulerLag}. (Variables for each physical term are omitted for simplicity.) The optimization variables consist of joint actuations $\boldsymbol{\tau}_t$ and the spring-mass model parameters $\mathbf{x}_t$. Specifically, the contact force model in Eq.~\ref{Eq:springmass} is written in a vector representation as $\boldsymbol{\lambda}_{p,t}=\mathbf{A}_{p,t}\mathbf{x}_{p,t}$, where $\mathbf{A}_{p,t}=s_{p,t} [-\mathbf{b}_{h,t},-\mathbf{b}_{n,t},-\mathbf{v}_{C,t}]$ include the position-related parameters, and $\mathbf{x}_{p,t}=[k_{h,t};k_{n,t};c_t]$ involve the unknown stiffness and damping parameters. Concatenating the contact forces of all modeled points, we have $\boldsymbol{\lambda}_t=\mathbf{A}_t\mathbf{x}_t$. Additionally, the linear inequality constraints on $\mathbf{x}_t$ are specified considering the maximal contact forces that can be experienced when a human is standing normally. The formulated optimization problem is a standard Quadratic Programming problem for which the global minimum is found by utilizing CVXOPT~\cite{diamond2016cvxpy}. The final solution yields forces that comply with the constraints set by the spring-mass model and optimally satisfy the Euler-Lagrange equations. Utilizing the inferred forces enables effective incorporation of physics into the model by imposing the following physics-inspired losses.

\noindent\textbf{Force Loss.} We first employ the derived motion forces to guide the model to produce realistic motion forces and extract meaningful latent representations for predicting physically plausible motion. Specifically, we introduce a linear layer to project the latent representation $\mathbf{E}_{latent}$ to motion forces $\{\boldsymbol{\lambda}_t,\boldsymbol{\tau}_t\}_{t=1}^{T}$ based on the contact force model. Given the derived forces $\bar{\boldsymbol{\tau}}_t$ and $\bar{\boldsymbol{\lambda}}_t$, we train the model by minimizing the mean absolute error as:
\begin{equation}
\begin{split}
\mathcal{L}_{force}=\sum_{t=1}^T\gamma_{\tau}\|\boldsymbol{\tau}_t-\bar{\boldsymbol{\tau}}_t\|_1+\gamma_{\lambda}\|\boldsymbol{\lambda}_t-\bar{\boldsymbol{\lambda}}_t\|_1.
\end{split}
\end{equation}
\noindent\textbf{Contact Loss.} Moreover, we apply constraints to the vertices experiencing contact forces, obtaining realistic contact behavior through the following contact loss:
\begin{equation}
\label{Eq:incontact}
    \mathcal{L}_{contact}=\sum_{t=1}^T\frac{1}{n_{C_t}}\sum_{n_i\in \mathcal{C}_t}
    \gamma_{v} \|\mathbf{v}_{n_i,t}\|_1 + \gamma_{z} |z_{n_i,t}|,
\end{equation}
where $\mathcal{C}_t$ denotes the set of vertices that exhibit contact forces (calculated via Eq.~\ref{Eq:pseudoforce}), and $n_{C_t}$ is the set size. $\mathbf{v}_{n_i,t}$ and $z_{n_i,t}$ represent the velocity and vertical distance to the ground of the $n_i^{th}$ vertex, respectively. Minimizing Eq.~\ref{Eq:incontact} encourages those vertices that experience large contact forces to have smaller velocity and be closer to the ground. 

\noindent\textbf{Euler-Lagrange Loss.} Furthermore, we incorporate a loss derived from the Euler-Lagrange equations to ensure the reconstructed motion adheres to the physics equations:
\begin{equation}
\label{Eq:euler}
\resizebox{.89\hsize}{!}{$
    \mathcal{L}_{euler} = \sum_{t=1}^T\|  \mathbf{M}_t\ddot{\mathbf{q}}_t +\mathbf{C}_t + \mathbf{g}_t - \mathbf{J}_{C,t}^T\bar{\boldsymbol{\lambda}}_t - \bar{\boldsymbol{\tau}}_t \|_1 
    $}.
\end{equation}
It is worth noting that all terms in the loss function are analytically computed and fully differentiable with respect to the model outputs thanks to the physics-based body model.

\noindent\textbf{Total Model Training Loss.} Combining all the physics-inspired losses with the reconstruction loss, we obtain the total training loss function: 
\begin{equation}
\label{Eq:totalloss}
    \mathcal{L} = \mathcal{L}_{recon} + \mathcal{L}_{force} + \mathcal{L}_{contact} + 
    \mathcal{L}_{euler}.
\end{equation}
We utilize Eq.~\ref{Eq:totalloss} to train the Transformer encoder-decoder backbone solely using motion capture data. Once the model is trained, it is directly added on top of the kinematics-based model to obtain improved motion estimates and infer motion forces, without the need of model fine-tuning.

%% file: sec/4_experiment.tex
\section{Experiment}

\noindent\textbf{Datasets.} During training, we use AMASS~\cite{mahmood2019amass}, a collection of motion capture datasets featuring a diverse range of subjects and actions. For evaluation, we utilize the test set of Human3.6M~\cite{ionescu2013human3} and 3DOH~\cite{zhang2020object}. Human3.6M encompasses common activities such as walking and sitting down. In contrast to certain physics-based methods that focus solely on sequences involving interactions with the ground, we adhere to the standard protocol and evaluate our method on all actions. 3DOH includes sequences of human-object interactions, such as opening a box --- representing a challenging testing setting with significant occlusions. Furthermore, we utilize PennAction~\cite{zhang2013actemes} to demonstrate that our approach helps improve human action recognition. PennAction comprises over 2K online videos of 15 sports actions, such as baseball pitching and bowling.

\noindent\textbf{Evaluation Metrics.} We evaluate 3D reconstruction error (\textit{Rec. Error}) and physical plausibility (\textit{Phys. Plausibility}). \textit{Rec. Error} includes the Mean Per-Joint Position Error (MJE in mm) and MJE after the Procrustes Alignment (P-MJE in mm). \textit{Phys. Plausibility} involves metrics introduced by prior methods~\cite{shimada2020physcap,kocabas2020vibe,yuan2021simpoe}, including: (1) the average difference between the predicted and the ground truth acceleration (acceleration error ACCL in mm/frame$^2$); (2) the difference between the predicted and the ground truth joint velocity magnitude (velocity error VEL in mm/frame); (3) the average displacement between two adjacent frames of those in-contact vertices (foot sliding FS in mm); (4) the average distance to the ground of those mesh vertices below the ground (ground penetration GP in mm).

\noindent\textbf{Implementation.} The Transformer backbone consists of standard encoder and decoder layers, with 6 layers, 8 attention heads, and 1024 embedding dimensions. The model's input sequence length is 16, aligning with most existing methods. For efficient Transformer training~\cite{vaswani2017attention}, we initially use the ground truth to extract the output embeddings for 20 epochs, followed by an additional 5 epochs using the prediction. We employ the Adam optimizer~\cite{kingma2014adam} with a weight decay of $10^{-4}$. The initial learning rate is $10^{-5}$ and decreases to 0.8 after every 5 epochs. The hyperparameters are empirically set as: $\gamma_{q}=2e^{3}$, $\gamma_{J}=1e^{5}$, $\gamma_{\tau}=5$, $\gamma_{\lambda}=1$, $\gamma_{v}=100$, and $\gamma_{z}=200$.

\begin{table}[t]
\tabcolsep=0.02in
    \begin{center}
    \scalebox{0.8}{
        \begin{tabular}{ l  cc  cc  cccc }
    \toprule
    \multirow[b]{2}{*}{Method} & \multirow[b]{2}{*}{\shortstack[c]{Physics\\Engine}} & \multirow[b]{2}{*}{\shortstack[c]{Video\\Label}} & \multicolumn{2}{c}{\textit{Rec. Error}} & \multicolumn{4}{c}{\textit{Phys. Plausibility}}  \\ \cmidrule{4-5} \cmidrule(lr){6-9}
    &  & & MJE & P-MJE & ACCL & VEL & FS & GP \\ \cmidrule{1-9}
    HybriK$^\dag$ \cite{li2021hybrik} & - & - & 55.4 & \textbf{33.6} & - & - & - & - \\
    *CLIFF \cite{li2022cliff} & - & - & \textbf{52.2} & 36.8 & 15.4 & 6.8 & 8.3 & 9.3
    \\ \cmidrule{1-9}
    VIBE \cite{kocabas2020vibe} & - & - & 61.3 & 43.1 & 15.2 & 25.5 & 15.1 & 12.6 \\
    *PoseBert \cite{baradel2022posebert} & - & - & 54.9 & 37.5 & 5.0 & 4.0 & 10.0 & 12.8  \\
    GLoT \cite{shen2023global} & - & - & 67.0 & 46.3 & 3.6 & - & - & - \\
    PMCE \cite{you2023co} & - & - & 53.5 & 37.7 & 3.1 & - & - & -
    \\\cmidrule{1-9}
    PhysCap \cite{shimada2020physcap} & Yes & - & 97.4 & 65.1 & - & 7.2 & - & - \\
    NeurPhys \cite{shimada2021neural} & Yes & Yes & 76.5 & 58.2 & - & 4.5 & - & - \\
    Xie \textit{et al.} \cite{xie2021physics} & Yes & - & 68.1 & - & - & 4.0 & - & -\\
    SimPoE \cite{yuan2021simpoe} & Yes & Yes & 56.7 & 41.6 & 6.7 & - & 3.4 & 1.6 \\
    NeurMoCon \cite{huang2022neural} & Yes & Yes & 72.5 & 54.6 & - & 3.8 & - & - \\ 
    TrajOpt \cite{gartner2022trajectory} & Yes & - & 84 & 56 & - & - & - & -  \\
    DiffPhy \cite{gartner2022differentiable} & Yes & - & 81.7 & 55.6 & - & - & - & - \\
    D\&D$^\dag$ \cite{li2022d} & No & Yes & 52.5 & 35.5 & 6.1 & - & 5.8 & \textbf{1.5} \\
    Huang \textit{et al.} \cite{ju2023physics} & Yes & Yes & 55.4 & 41.3 & - & 3.5 & - & - 
    \\ \cmidrule{1-9} 
    \multirow{2}{*}{\textbf{{\ours}} (\textbf{Ours})} &\multirow{2}{*}{\textbf{No}} &\multirow{2}{*}{\textbf{No}} & \multirow{2}{*}{52.7} & \multirow{2}{*}{36.7} & \textbf{2.5} & \textbf{3.4} & \textbf{2.6} & \textbf{1.5} \\
    & & & & &{\color{ForestGreen}$\downarrow$83.8} & {\color{ForestGreen}$\downarrow$50.0} & {\color{ForestGreen}$\downarrow$68.7} & {\color{ForestGreen}$\downarrow$83.9} \\
    \bottomrule
    \end{tabular}}
    \end{center}
    \vspace{-0.5cm}
    \caption{\textbf{Evaluation on Human3.6M.} Methods in the top block use image inputs, those in the middle use video inputs, and those in the bottom are physics-based. Current physics-based methods require 3D annotated videos (``Video Label") for training or adopt a optimization-based framework. Methods marked by $^\dag$ are evaluated on 3D joints computed from fitted body models~\cite{loper2014mosh} instead of the one provided in the original datasets. For those marked ``*", the results are from their officially released models. All other results are taken from the respective papers. Evaluation of PoseBert and {\ours} is based on CLIFF. The green numbers represent percentages of the relative improvement of our approach over CLIFF. For all metrics, smaller values are preferred.}
    \label{tab:quantitativesota}
\end{table}

\subsection{Comparison with State-of-the-Arts (SOTAs)}
\label{sec:sota}

\noindent\textbf{Improvements to Kinematics-based Methods.} As seen in Tab.~\ref{tab:quantitativesota},  {\ours}, significantly improves the physical plausibility of kinematics-based motion estimates. Whether they take images or video frames as input, the kinematics-based methods often struggle with physical plausibility. For example, CLIFF retains competitive per-frame reconstruction accuracy but provides poor performance on all the physical plausibility evaluation metrics. Applying {\ours} to CLIFF significantly enhances its physics plausibility. Notably, the acceleration error (ACCL) and foot skating (FS) are reduced by 83.8\% and 68.7\% respectively. Like {\ours}, PoseBert leverages a Transformer-encoder pre-trained on 3D motion capture data, but it does not consider physics. PoseBert reduces ACCL and VEL somewhat, but unlike {\ours} it fails to decrease the foot skating and ground penetration error. In Supp.~\ref{supp:differentbackbones}, we demonstrate that {\ours} also improves other kinematics-based models besides CLIFF (SPIN~\cite{kolotouros2019learning} and IPMAN~\cite{tripathi20233d}) and the improvements are consistently more significant than PoseBert.

\noindent\textbf{Advantages over Physics-based Methods.} {\ours} surpasses existing physics-based methods without relying on physics engines or 3D annotated videos for training. As illustrated in Tab.~\ref{tab:quantitativesota}, existing physics-based methods generally exhibit improved physical plausibility compared to kinematics-based methods. They typically employ a physics engine separate from their learning models to compute physical parameters and simulate body motion. They require 3D annotations paired with input videos for training or are confined to optimization-based approaches. In contrast, our approach avoids the need for 3D annotated videos by exploiting an innovative self-supervised learning framework. We bridge the gap between body kinematics and physics through a physics-based body representation and contact force model, allowing the seamless integration of physics with deep models. Comparing the performance, our model achieves competitive reconstruction accuracy with more significant advancements in physical plausibility (Tab.~\ref{tab:quantitativesota}). For instance, although all other methods utilize training data from Human3.6M, our approach yields an acceleration error that is 2.4 times less than that of its nearest competitor D\&D (2.5 vs. 6.1 mm/frame$^2$), and foot skating that is 76\% of second-best SimPoE's (2.6mm vs. 3.4mm). In Supp.~\ref{supp:evalglobalmotion}, we demonstrate that our approach produces improved global motion recovery as well.

\noindent\textbf{Robustness under Occlusion.} Our approach is robust to occlusion, as demonstrated through the evaluation on 3DOH (Tab.~\ref{tab:3doh}). Despite the significant occlusions and complex human-object interaction motions included in 3DOH, applying {\ours} to CLIFF produces consistent improvement on motion estimates. The final model performance outperforms existing physics-based methods. For example, our approach achieves a velocity error of 6.5 mm/frame (Tab.~\ref{tab:3doh}, VEL), 45.8\% less than CLIFF's 12.0, and 27.0\% less than the 8.9 attained by SOTA Huang \textit{et al.}. Furthermore, on 3DOH, our approach surpasses existing physics-based methods in reconstruction accuracy by a large margin. Specifically, our approach achieves P-MJE of 33.5, 54.0\% less than Huang \textit{et al.}'s 72.8. Existing physics-based methods do not utilize 3DOH for training, and their performance degrades on new testing sequences. In contrast, our approach maintains better generalization ability and effectively leverages the favourable per-frame 3D body reconstruction of kinematics-based estimates to generate accurate and physical plausible motion.

\begin{table}[t]
\tabcolsep=0.04in
    \begin{center}
    \scalebox{0.9}{
\begin{tabular}{ l cc  cccc }
    \toprule
    \multirow[b]{2}{*}{Method} & \multicolumn{2}{c}{\textit{Rec. Error}} & \multicolumn{4}{c}{\textit{Phys. Plausibility}}  \\  \cmidrule{2-3} \cmidrule(lr){4-7}
     & MJE & P-MJE & ACCL & VEL & FS & GP\\ 
     \cmidrule{1-7}
    *CLIFF \cite{li2022cliff} & \textbf{53.0} & 34.4 & 26.0 & 12.0 & 10.8 & 12.6 \\
    \cmidrule{1-7} 
    VIBE \cite{kocabas2020vibe} & 98.1 & 61.8 & - & 26.5 & - &- \\
    *PoseBert \cite{baradel2022posebert} & 54.8 & 34.1 & 6.6 & 6.9 & 14.0 & 10.4  \\
    \cmidrule{1-7} 
    NeurPhys \cite{shimada2021neural} & 107.8 & 93.3 & - & 12.2 & - & - \\
    NeurMoCon \cite{huang2022neural} & 93.4 & 86.7 & - & 9.2 & - & - \\ 
    Huang \textit{et al.} \cite{ju2023physics} & 79.3 & 72.8 & - & 8.9 & - & - \\
    \cmidrule{1-7}
    \multirow{2}{*}{\textbf{{\ours}} (\textbf{Ours})} & \multirow{2}{*}{\textbf{53.0}} & \multirow{2}{*}{\textbf{33.3}} & \textbf{4.6} & \textbf{6.5} & \textbf{4.7} & \textbf{0.1} \\
    & & & {\color{ForestGreen}$\downarrow$82.3} & {\color{ForestGreen}$\downarrow$45.8} & {\color{ForestGreen}$\downarrow$56.5} & {\color{ForestGreen}$\downarrow$99.2} \\
    \bottomrule
    \end{tabular}}
    \end{center}
    \vspace{-0.5cm} \caption{\textbf{Evaluation on 3DOH.} Evaluation of PoseBert and {\ours} is based on CLIFF. The green numbers represent percentages of the relative improvement of our approach over CLIFF.}
    \label{tab:3doh}
\end{table}

\subsection{Ablation Study}
\label{sec:ablation}

\begin{table}[t]
\tabcolsep=0.02in
    \begin{center}
    \scalebox{0.9}{
    \begin{tabular}{ cccc  cc  cccc }
    \toprule
    \multicolumn{4}{c}{Training Losses} & \multicolumn{2}{c}{\textit{Rec. Error}} & \multicolumn{4}{c}{\textit{Phys. Plausibility}}  \\ \cmidrule(lr){1-4} \cmidrule{5-6} \cmidrule(lr){7-10}
    $\mathcal{L}_{recon}$ & $\mathcal{L}_{force}$ & $\mathcal{L}_{contact}$ & $\mathcal{L}_{euler}$ & MJE & P-MJE & ACCL & VEL & FS & GP \\ \cmidrule{1-10}
    - & - & - & - & 52.2 & 36.8 & 15.4 & 6.8 & 8.3 & 9.3 \\ \cmidrule{1-10}
    $\checkmark$ &  & & & 52.7 & 36.7 & 2.5 & 3.5 & 7.1 & 6.9 \\
    $\checkmark$ & $\checkmark$ & & & 52.7 & 36.7 & 2.5 & 3.4 & 6.5 & 5.6 \\
    $\checkmark$ & $\checkmark$ & $\checkmark$ & & 53.0 & 36.8 & 2.5 & 3.4 & 4.1 & 1.7 \\
    \cmidrule{1-10}
    $\checkmark$ & $\checkmark$ & $\checkmark$ & $\checkmark$ & 52.7 & 36.7 & 2.5 & 3.4 & 2.6 & 1.5 \\
    \bottomrule
    \end{tabular}}
    \end{center}
    \vspace{-0.5cm}
    \caption{\textbf{Ablation on the Training Losses.} The evaluation is on Human3.6M. The first row denotes the kinematics-based model}
    \label{tab:ablation}
\end{table}

\begin{figure}[t]
    \centering
    \includegraphics[width=0.47\textwidth]{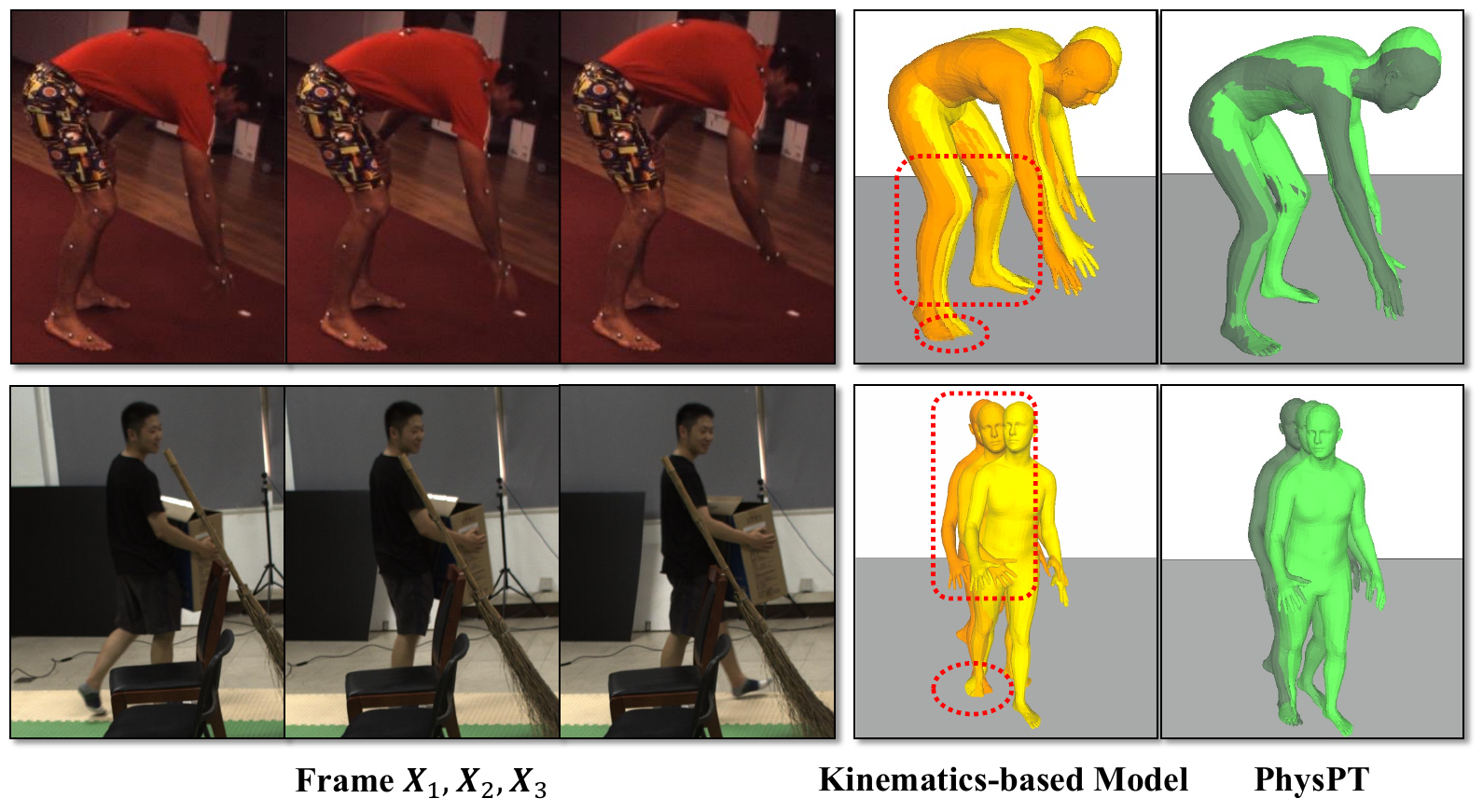}
    \vspace{-0.25cm}
    \caption{\textbf{Qualitative Evaluation on Utilizing PhysPT.} The body color of each figure represents the reconstruction at different time frames (lighter colors indicate later time frames). Ground penetration and motion jittering exhibited in the reconstructed motion are marked by red circle and rectangle, respectively.}
    \label{fig:abalationquality}
\end{figure}

\noindent\textbf{Effectiveness of {\ours}.} We first study the effectiveness of the Transformer encoder-decoder backbone and the physics-inspired training losses. According to the evaluation results detailed in Tab.~\ref{tab:ablation}, when trained exclusively with the reconstruction loss (Eq.~\ref{eq:recon}), the model maintains the reconstruction accuracy while reducing the acceleration and velocity errors of the kinematics-based estimates (Tab.~\ref{tab:ablation}-row2 over row1). The reduction is more significant than that observed in PoseBert (Tab.~\ref{tab:quantitativesota}), demonstrating the advantages of leveraging the Transformer encoder-decoder rather than using the encoder solely. However, the foot skating and ground penetration errors are reduced to a much lesser extent. Reducing them significantly requires further imposing the physics-inspired losses. Specifically, when force labels are used for training, the foot skating error drops from 7.1mm to 6.5mm and the ground penetration error drops from 6.9mm to 5.6mm. Imposing the contact loss (Eq.~\ref{Eq:incontact}) further reduces the errors but sacrifices the reconstruction accuracy (Tab.~\ref{tab:ablation}-row3). To obtain the best model, {\ours}, we leverage all the physics-inspired losses for training. In Fig.~\ref{fig:abalationquality}, we showcase that utilizing {\ours} effectively reduces the motion jitter and foot penetration exhibited by kinematics-based estimates. For example, the kinematics-based model can produce excessive motion jitter of the lower body even when only the upper body moves (Fig.~\ref{fig:abalationquality}-row1) or be affected by occlusion in the input video frame (Fig.~\ref{fig:abalationquality}-row2). By contrast, {\ours} resolves these issues by integrating physics with the Transformer.

\begin{figure}[t]
    \centering
    \includegraphics[width=0.47\textwidth]{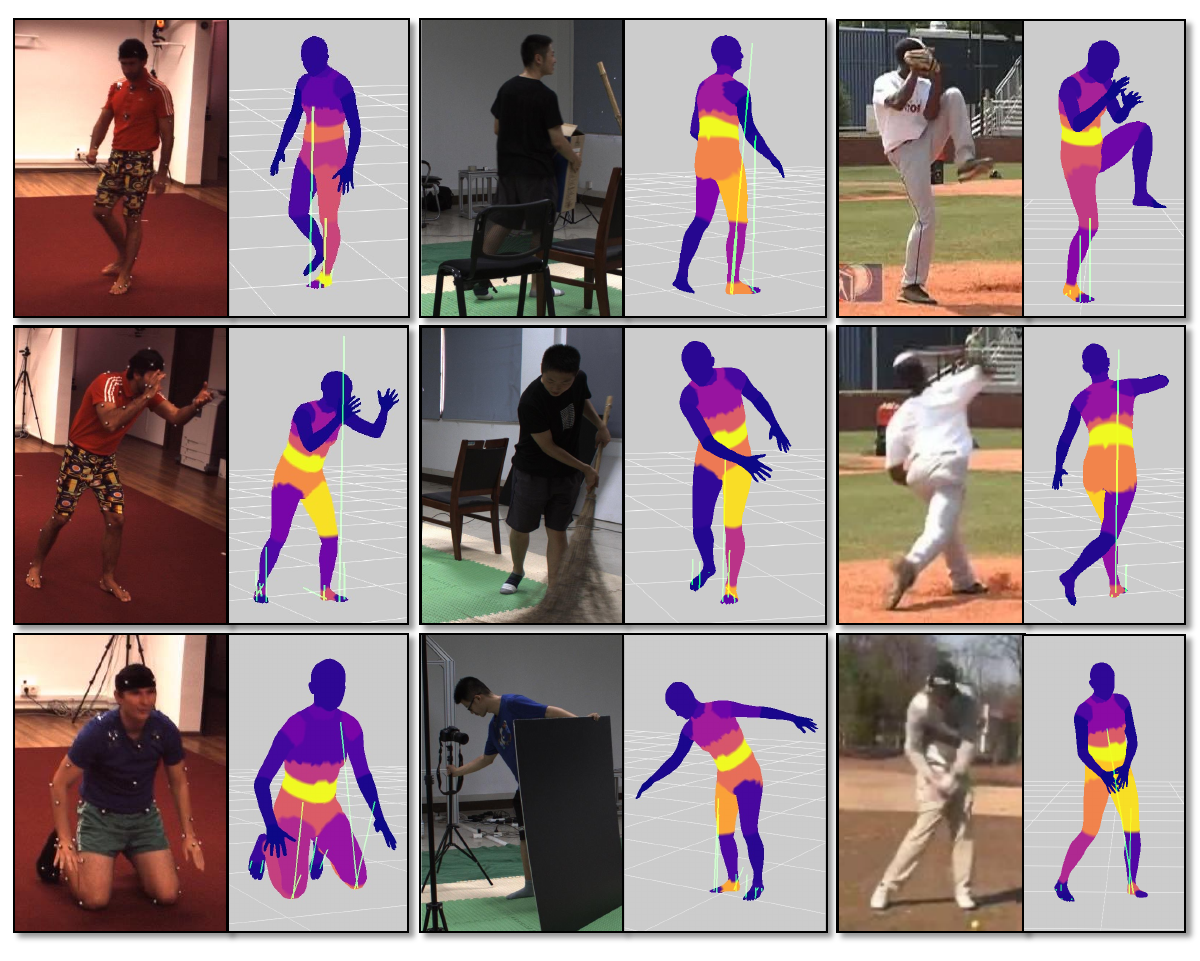}
    \vspace{-0.25cm}
    \caption{\textbf{Qualitative Evaluation with Force Estimation Visualization.} The testing image frames are from Human3.6M (left), 3DOH (middle), and PennAction (right).}
    \label{fig:motionandforcce}
\end{figure}

\noindent\textbf{Motion Reconstruction with Force Estimation.} Our approach can generate accurate 3D motion and infer forces, as illustrated qualitatively in Fig.~\ref{fig:motionandforcce}. The inferred forces offer valuable insights into body dynamic behavior. For instance, in the left column of Fig.~\ref{fig:motionandforcce}, significant contact forces and joint actuations are evident on the left foot when the subject walks forward with the left foot (top) or on the left leg and when the body leans forward to the left (middle). The estimated forces also capture the contact behavior of various body parts, such as the knee (bottom). Moreover, our approach effectively handles occlusion (Fig.~\ref{fig:motionandforcce}, 3DOH) and is applicable to in-the-wild videos (Fig.~\ref{fig:motionandforcce}, PennAction).

\subsection{Improvements to Human Action Recognition} 
\label{sec:actionrecogntion}

In the preceding sections, we illustrate that {\ours} generates more physically-realistic motion and produces reasonable force estimates. The improved motion and additional force estimates can successfully improve downstream tasks such as human action recognition. We demonstrate this through the evaluation of human action recognition on PennAction. Specifically, we employ a recent skeleton-based recognition model proposed by \cite{chen2021channel}. We utilize the motion and forces generated by our approach as model input and evaluate the corresponding recognition accuracy. We summarize the evaluation results with comparison to recent skeleton-based recognition models in Tab.~\ref{tab:humanaction}. In this comparison, we exclude the physics-based recognition model discussed in the related work (\cite{mansur2012inverse}) as it relies on 3D motion capture data and is not adept at handling in-the-wild videos. Conventional methods primarily rely on 2D body pose as model input. In contrast, our approach excels by leveraging 3D body pose information. Particularly, as shown in Tab.~\ref{tab:humanaction}, utilizing the motion generated by {\ours} yields better performance compared to using kinematics-based estimates (96.8\% over 96.0\% in accuracy). Note that, using estimated forces alone, the accuracy is superior to that of existing methods (94.4\% over UNIK's 94.0\%), further demonstrating the effectiveness of our approach in estimating human dynamics. Finally, the combination of motion and force estimates leads to a significant performance boost, achieving the best recognition accuracy of 98.0\%. In Supp.~\ref{supp:actionrecognition}, we provide action-wise evaluation, illustrating that utilizing forces enhances performance, particularly in cases where relying solely on 3D joint positions falls short, such as when different actions have similar body movement patterns.

\begin{table}[t]
\tabcolsep=0.02in
    \begin{center}
        \begin{tabular}{ c | cc | cccc }
    \toprule
    \multirow[b]{2}{*}{Method} & \multirow[b]{2}{*}{\shortstack[c]{HDM \\ \cite{zhao2019bayesian}}} & \multirow[b]{2}{*}{\shortstack[c]{UNIK \\\cite{yang2021unik}}} & \multicolumn{4}{c}{\textbf{Ours}}  \\ 
    \cmidrule(lr){4-7}
     &  &  & $\mathbf{J}_{kin}$ & $\mathbf{J}_{phys}$ & $\mathbf{F}$ & $\mathbf{J}_{phys}$+$\mathbf{F}$ \\ \midrule 
    Top-1 Acc. ($\uparrow$) & 93.4 & 94.0 & 96.0 & 96.8 & 94.4 & \textbf{98.0} \\
    \bottomrule
    \end{tabular}
    \end{center}
    \vspace{-0.5cm}
    \caption{\textbf{Human Action Recognition.} The evaluation is on PennAction. We report the Top-1 accuracy in percentage. $\mathbf{J}_{kin}$ and $\mathbf{J}_{phys}$ stands for the 3D body joint positions estimated by the kinematics-based model and {\ours}, respectively. $\mathbf{F}$ indicates the motion forces output by {\ours}.}
    \label{tab:humanaction}
\end{table}

%% file: sec/5_conclusion.tex
\section{Conclusion} 

In summary, we describe {\ours} (Physics-aware Pretrained Transformer), which generates more physically plausible motion estimates than previous methods and infers motion forces. {\ours} exploits a Transformer encoder-decoder backbone trained through self-supervised learning and it integrates physics principles. Specifically, we craft a physics-based body representation and a continuous contact force model. We introduce novel physics-inspired training losses. Leveraging them for model training enables {\ours} to effectively capture physical properties of the human body and the forces it experiences. Through extensive experiments, we demonstrate the direct applicability of {\ours} to kinematics-based estimates results in the reconstruction of more physically-realistic motion and the inference of motion forces from monocular videos. Notably, for the first time, we demonstrate that these more accurate estimates of motion and force translate to improvements in an important downstream task: human action recognition.

\paragraph{Acknowledgement} This work is supported in part by the Rensselaer-IBM AI Research Collaboration (http://airc.rpi.edu), part of the IBM AI Horizons Network.

%% file: sec/X_suppl.tex
\clearpage
\maketitlesupplementary
\setcounter{page}{1}
\setcounter{section}{0}
\appendix

In this supplementary material, we first provide additional details of our proposed approach:
\begin{itemize}[leftmargin=0.2in]
    \item Section~\ref{supp:globaltraj}: Global Trajectory Estimation
    \item Section~\ref{supp:physicsmodel}: Creation of Phys-SMPL, which includes (1) the Calculation of Body Part Volume, Mass, and Inertia Tensor; and (2) the Derivation of the Physical Parameters in the Euler-Lagrange Equations
    \item Section~\ref{supp:contactregion}: Selection of Body Contact Regions
\end{itemize}
We then present additional evaluation results:
\begin{itemize}[leftmargin=0.2in]
    \item Section~\ref{supp:differentbackbones}: Improvements over Different Kinematics-based 3D Body Reconstruction Models
    \item Section~\ref{supp:evalglobalmotion}: Evaluation on Global Motion Recovery
    \item Section~\ref{supp:pseudoforce}: Quality of the Generated Force Labels
    \item Section~\ref{supp:actionrecognition}: Action-wise Recognition Performance
\end{itemize}

\section{Global Trajectory Estimation}
\label{supp:globaltraj}

\begin{figure}[t]
    \centering
    \includegraphics[width=0.45\textwidth]{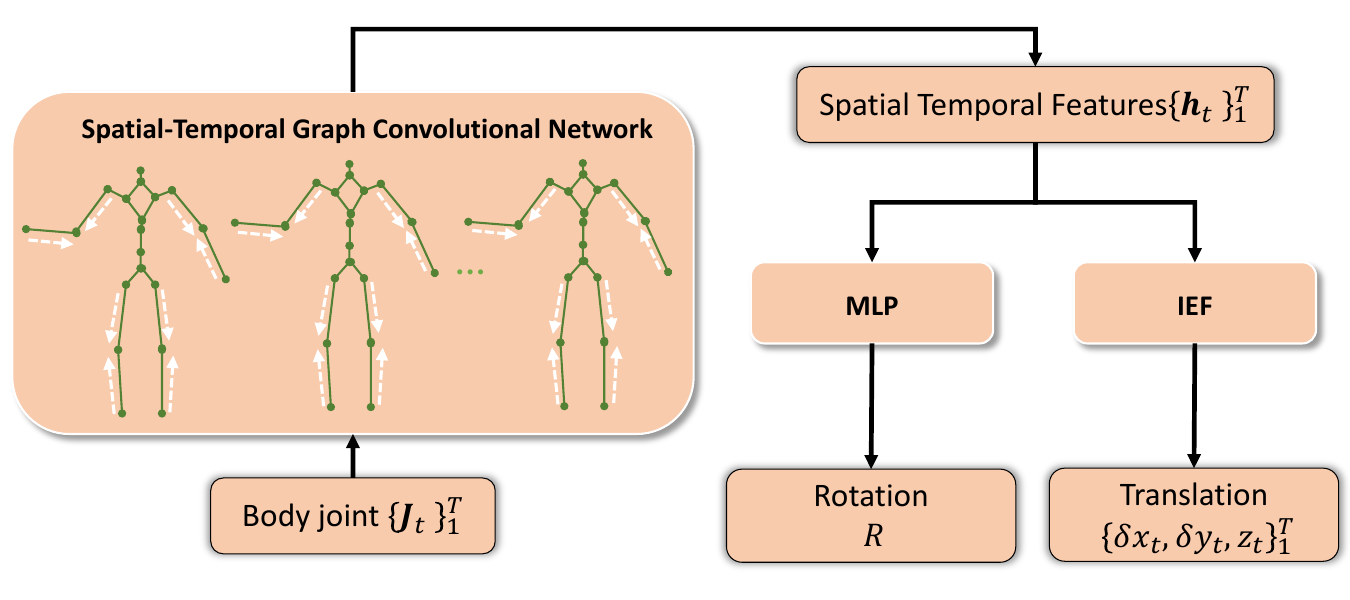}
    \caption{\textbf{Global Trajectory Estimation Model.} ``MLP" and ``IEF" represents Multi-layer Perceptrons and Iterative Error Feedback regression model~\cite{carreira2016human}, respectively.}
    \label{fig:globaltraj}
\end{figure}

Traditional image or video-based 3D human body reconstruction models provide estimates of 3D body configuration in the body frame. Additionally, they estimate a root rotation that transforms the estimated 3D configuration from the body frame to the camera frame. To effectively model human dynamics, a human body motion trajectory represented in the world frame is needed. Besides local body movements, the global motion trajectory further involves the relative rotation between the camera frame and the world frame, and the body translation in the world frame. Inspired by the framework proposed by \cite{yuan2022glamr}, we estimate these information through a global trajectory predictor with the model architecture illustrated in Figure~\ref{fig:globaltraj}. The input of the model are 3D body joint positions represented in the body frame $\{\mathbf{J}_{t}\}_{t=1}^T$. A Spatial-Temporal Graph Convolutional Networks (ST-GCN)~\cite{yan2018spatial} are then employed to extract spatial-temporal features for every time frame as:
\begin{equation}
    \{\mathbf{h}_{t}\}_{t=1}^T = \texttt{GCN}(\{\mathbf{J}_{t}\}_{t=1}^T),
\end{equation}
where $\mathbf{h}^{t}\in \mathbb{R}^{256}$. Typically, the camera pose remains constant throughout a video. The model thus estimates a single rotation matrix for all frames. Specifically, the extracted features $ \{\mathbf{h}_{t}\}_{t=1}^T$ are concatenated together and then input to a multi-layer perceptrons (MLP). The MLP module includes three fully connected layers with the ReLU activation functions~\cite{nair2010rectified} to predict the rotation transformation between the camera and world frame as:
\begin{equation}
    \mathbf{R}_c = \texttt{MLP}(\{\mathbf{h}_{t}\}_{t=1}^T).
\end{equation}
Combining the estimated $\mathbf{R}_c$ with the root rotation generated by the kinematics-based 3D human body reconstruction model at each frame leads to the global rotation $\{\mathbf{R}_{t}\}_{t=1}^T$ defined in Eq.~\ref{eq:generalizedposition}. Note that the prediction of the camera rotation can be readily extended to predicting rotations for every frame considering a moving camera. Furthermore, to estimate the global translation, we employ a regression model with Iterative Error Feedback (IEF)~\cite{carreira2016human} with 3 iterations. The employed regression model takes as input the exacted spatial-temporal features and outputs the 3D root joint positions represented in the world frame as:
\begin{equation}
    \delta_{x,t}, \delta_{y,t}, z_t = \texttt{IEF}(\mathbf{h}_{t}),
\end{equation}
where $\delta_{x,t}$ and $\delta_{y,t}$ are position changes in the horizontal directions at time frame $t$, and $z_t$ is the corresponding vertical position relative to the ground plane. Adding the position changes $(\delta_{x,t},\delta_{y,t})$ over time and combining them with the vertical position $z_t$ at different time frames, we obtain the final global translation $\{\mathbf{T}_{t}\}_{t=1}^T$ required to specify the generalized positions defined in Eq.~\ref{eq:generalizedposition}.

For training of the global trajectory predictor, we use AMASS~\cite{mahmood2019amass}. The training loss consists of the mean square errors between the predicted and the ground truth rotation and translation. The 3D motion sequences in AMASS are 3D trajectories represented in the world frame and only vary in the body translation. We hence introduce random rotation changes to the input to allow the model to predict the rotation changes. Meanwhile, we add random Gaussian noise to the input 3D joint positions to improve the model robustness. During training, we utilize the Adam optimizer~\cite{kingma2014adam} with a weight decay of $10^{-4}$. The initial learning rate is $10^{-3}$ and decreases to its 0.95 after every 15,000 steps. The total number of training epochs is 20. Note that the training of the global trajectory predictor is independent to certain 3D human body reconstruction models. Once the model is trained, we directly combine it with the 3D reconstruction model to generate the initial generalize positions $\{\hat{\mathbf{q}}_t\}_{t=1}^{T}$.

\section{Creation of Phys-SMPL}
\label{supp:physicsmodel}

Phys-SMPL characterizes the necessary physical properties of the human body required in modeling human dynamics through the Euler-Lagrange equations. Specifically, these physical information includes body mass and inertia of different body parts and is utilized in computing the physical terms in the Euler-Lagrange equations. In this section, we first introduce the way of calculating the physics information based on the geometry information provided by SMPL. Then, we present the analytical equations for computing the physical parameters in the Euler-Lagrange Equations.

\noindent\textbf{Calculation of Body Part Volume, Mass, and Inertia Tensor.} The original SMPL builds upon 3D triangle mesh models. The 3D triangle mesh model captures the body geometry information but not physics. To effectively model physical properties from the geometry information, we first compute the volume of different body parts. Specifically, we build a close mesh for each body part by closing the mesh along the boundary. Then, each mesh triangle in a body part combined with the body part center can form a 3D tetrahedron, the volume of which can be easily computed. Meanwhile, for each body part, its volume can be computed as the sum of the volumes of all the 3D tetrahedra belonging to that body part. Defining the origin of a body part as its geometric center, the volume of body part $i$ can thus be computed as:
\begin{equation}
        V_i = \sum_{j = 1}^{n_j} |\det(\mathbf{P}_{i,j,1},\mathbf{P}_{i,j,2},\mathbf{P}_{i,j,3})|,
\end{equation}
where $n_{j}$ is the total number of mesh triangles included in the $i^{th}$ body part, $\mathbf{P}_{i,j,1}$, $\mathbf{P}_{i,j,2}$, and $\mathbf{P}_{i,j,3}$ are the 3D vertex position of the $j^{th}$ triangle.

Using the computed volume, we can then determine the mass of each body part. Specifically, for the mean shape of SMPL, we consider it has a total body mass of 70 kg following the typical setting~\cite{xie2021physics}. We compute the mass of each body part by distributing the total body mass based on the average human body weight distribution~\cite{plagenhoef1983anatomical}. For subjects with different body shapes, we calculate their body mass based on the proportion of their body part volume relative to the mean shape. As human tissue exhibits varying mass density, instead of scaling the mass solely based on the body part volume, we consider the density differences between bone, muscle, and fat to introduce additional predefined scaling factors for different body parts following the study shown in ~\cite{prior2001muscularity}.

For the inertia tensor required to specify the Euler-Lagrange equations, we need to compute the inertia tensor for each body part relative to its root joint represented in the body part frame. In the following, we present the analytical equations to compute the inertia tensor $\mathbf{I}_i$ of the $i^{th}$ body part without loss of the generality. Firstly, we consider a body part is a rigid solid body that has a uniform mass density. The entries in the inertia tensor can be denoted as
\begin{equation}
    \mathbf{I}_i = \begin{bmatrix}
        I_{xx} & I_{xy} & I_{xz} \\ 
        I_{yx} & I_{yy} & I_{yz} \\ 
        I_{zx} & I_{zy} & I_{zz} \\
    \end{bmatrix}.
\end{equation}
Following the standard way to compute the inertia tensor~\cite{spong2008robot}, an entry in $\mathbf{I}_i$ is computed as 
\begin{equation}
    \frac{m_i}{V_i}\iiint_{(x,y,z)\in\mathbf{S}_{i}} f(x,y,z) dxdydz,
\end{equation}
where $m_i$, $V_i$, and $\mathbf{S}_{i}$ denotes the mass, volume, 3D integral region of the body part, respectively. Then, 
\begin{align}
    f(x,y,z) = \begin{cases}
        y^2 + z^2 & \text{for $I_{xx}$} \\
        x^2 + z^2 & \text{for $I_{yy}$} \\
        x^2 + y^2 & \text{for $I_{zz}$} \\
        -xy & \text{for $I_{xy}$ and $I_{yx}$} \\
        -xz & \text{for $I_{xz}$ and $I_{xz}$} \\
        -yz & \text{for $I_{yz}$ and $I_{yz}$} \\
\end{cases}.
\end{align}
As a body part is represented as a triangulate mesh, the integral can be computed as the sum of the integral computed within each tetrahedron of the body part as
\begin{equation}
    \frac{m_i}{V_i}\sum_{j=1}^{n_{i}} \iiint_{(x,y,z)\in\mathbf{S}_{i,j}} f(x,y,z) dxdydz,
\end{equation}
where $\mathbf{S}_{i,j}$ is the 3D integral region of the $j^{th}$ tetrahedron, and $n_{j}$ is the total number of mesh triangles of the $i$ body part. Given that each body part is assumed to have a uniform mass density, the Center of Mass (COM) coincides with the centroid of that body part. As shown in \cite{kallay2006computing}, the integral over the $j^{th}$ tetrahedron of the $i^{th}$ body part can be computed as 
\begin{equation}
\resizebox{.95\hsize}{!}{$
    \frac{v_{i,j}}{20} (f(\mathbf{P}_{i,j,1})+f(\mathbf{P}_{i,j,2})+f(\mathbf{P}_{i,j,3})+f(\mathbf{P}_{i,j,1}+\mathbf{P}_{i,j,2}+\mathbf{P}_{i,j,3})),
    $}
\end{equation}
where $\mathbf{P}_{i,j,1}$, $\mathbf{P}_{i,j,2}$, and $\mathbf{P}_{i,j,3}$ represent the 3D vertex positions of the $j^{th}$ tetrahedron in the body frame with regards to the COM. 

As introduced above, the calculation of the body mass and inertia tensor is based on the study of human anatomy and is accomplished using SMPL's geometry information. This approach avoids the use of unrealistic proxy bodies and establishes a direct mapping between the physics information and the shape parameters.

\noindent\textbf{Derivation of the Physical Parameters in the Euler-Lagrange Equations.} The physical parameters in the Euler-Lagrange equations are functions of the generalized positions, as well as the body mass and inertia. In this section, we present analytical equations to compute these physical parameters, including the contact Jacobian $\mathbf{J}_C$, the generalized mass matrix $\mathbf{M}\big(\mathbf{q} ;\ \mathbf{m},\mathbf{I}\big)$, the gravitational force $\mathbf{g}\big(\mathbf{q};\mathbf{m},\mathbf{I}\big)$, and the generalized bias force $\mathbf{C}\big(\mathbf{q},\dot{\mathbf{q}};\ \mathbf{m},\mathbf{I}\big)$.

For the contact Jacobian, it is the Jacobian matrix that maps a contact force represented in the Cartesian coordinates to the generalized coordinates. The contact Jacobian matrix only relates to the position of a contact point. For the human body, it can receive multiple contact forces. Below we derive the contact Jacobian matrix applied to one contact position $C$ without loss of generality. 

As shown in \cite{RobotDynamics}, the contact Jacobian can be computed as,
\begin{equation}
    \mathbf{J}_{C} = \begin{bmatrix}
        \mathbf{e}_x & \mathbf{e}_y &
        \mathbf{e}_z &
        \boldsymbol{\xi}_i & \cdots
    \end{bmatrix},
\end{equation}
where $\mathbf{e}_x$, $\mathbf{e}_y$, and $\mathbf{e}_z$ are the unit vectors along the $x$, $y$, and $z$ directions, respectively. $\mathbf{e}_x$, $\mathbf{e}_y$, and $\mathbf{e}_z$ correspond to the global translation defined in the generalized position $\mathbf{q}$. For the other columns corresponding to joint rotations, they are computed as
\begin{equation}
    \boldsymbol{\xi}_i = \begin{cases}
    _\mathcal{A}\mathbf{h}_i \times _\mathcal{A}\mathbf{r}_{iC}, & \text{if $i$ is a parent joint},\\
    \mathbf{0}, & \text{otherwise}.\\
    \end{cases}
\end{equation}
where ${}_\mathcal{A}\mathbf{h}_i$ is the rotation axis of $\mathbf{q}_i$ represented in the world frame $\mathcal{A}$, and is computed as 
\begin{equation}
    {}_\mathcal{A}\mathbf{h}_i = \mathbf{R}_{0i-1} {}_\mathcal{B}\mathbf{h}_i.
\end{equation}
$\mathbf{R}_{0i-1}$ is the rotation matrix describing the transformation from the body part frame to the world frame. Moreover, ${}_\mathcal{A}\mathbf{r}_{iC}$ is the 3D position of the contact point relative to its root joint $i$ represented in the world frame. ${}_\mathcal{A}\mathbf{r}_{iC}$ is computed as 
\begin{equation}
    {}_\mathcal{A}\mathbf{r}_{iC} = \mathbf{r}_{C}-\mathbf{P}_{i}, 
\end{equation}
where $\mathbf{r}_{C}$ and $\mathbf{P}_{i}$ are 3D positions of the contact point and the root joint represented in the world frame, respectively.

For the generalized mass matrix $\mathbf{M}\big(\mathbf{q};\ \mathbf{m},\mathbf{I}\big)$, it is a function of the body part mass $\mathbf{m}$, inertia $\mathbf{I}$, and the generalized position $\mathbf{q}$. The generalized mass matrix can be computed as the sum of the generalized mass of each individual body part as:
\begin{equation}
    \mathbf{M}\big(\mathbf{q};\ \mathbf{m},\mathbf{I}\big) = \sum_{n=1}^{24} \mathbf{J}_{S,n}^T m_n \mathbf{J}_{S,n} + \mathbf{J}_{R,n}^T \mathbf{R}_{0,n}\mathbf{I}_n \mathbf{R}_{0,n}^T\mathbf{J}_{R,n}
\end{equation}
where $\mathbf{J}_{S,n}$ is the Jacobian matrix computed in the world frame. The calculation of $\mathbf{J}_{S,n}$ is similar to the contact Jacobian matrix but considers the root joint of a body part as the target position to computed the Jacobian matrix. Moreover, $\mathbf{R}_{0,n}$ is the pose of the $n^{th}$ body part, and $\mathbf{J}_{R,n}$ is the angular Jacobian computed as
\begin{equation}
    \mathbf{J}_{R,n} = \begin{bmatrix}
        \mathbf{0} & \mathbf{0} & \mathbf{0} & 
        \boldsymbol{\xi}_i & \cdots & \mathbf{0} 
    \end{bmatrix},
\end{equation}
where 
\begin{equation}
    \boldsymbol{\xi}_i = \begin{cases}
    _\mathcal{A}\mathbf{h}_i , & \text{if $i$ is a parent joint},\\
    \mathbf{0}, & \text{otherwise}.\\
    \end{cases}
\end{equation}

For the gravitational force term, it is computed as 
\begin{equation}
    \mathbf{g}\big(\mathbf{q};\ \mathbf{m},\mathbf{I}\big) = -\sum_{n=1}^{24} \mathbf{J}_{S,n}^T m_n\mathbf{g},
\end{equation}
where $\mathbf{J}_{S,n}$ is the Jacobian matrix of the $n^{th}$ body part and $\mathbf{g}$ is the gravitational acceleration.

For the generalized bias force $\mathbf{C}\big(\mathbf{q},\dot{\mathbf{q}};\ \mathbf{m},\mathbf{I}\big)$, it is computed as
\begin{align}
    \mathbf{C}\big(\mathbf{q},\dot{\mathbf{q}};\ \mathbf{m},\mathbf{I}\big) & = \sum_{n=1}^{24} \mathbf{J}_{S,n}^T m_n \Dot{\mathbf{J}}_{S,n} \Dot{\mathbf{q}} \\
    &+ \mathbf{J}_{R,n}^T (\mathbf{R}_{0,n}\mathbf{I}_n \mathbf{R}_{0,n}^T \Dot{\mathbf{J}}_{R,n} \Dot{\mathbf{q}} \\
    &+ \mathbf{J}_{R,n} \Dot{\mathbf{q}} \times \mathbf{R}_{0,n}\mathbf{I}_n \mathbf{R}_{0,n}^T \mathbf{J}_{R,n} \Dot{\mathbf{q}}),
\end{align}
where $\Dot{\mathbf{J}}_{S,n}$ and $\Dot{\mathbf{J}}_{R,n}$ are the time derivatives of the linear and angular Jacobian. Specifically,
\begin{equation}
    \Dot{\mathbf{J}}_{S,n} = \begin{bmatrix}
        \mathbf{0} & \mathbf{0} &
        \mathbf{0} &
        \Dot{\boldsymbol{\xi}}_i & \cdots
    \end{bmatrix}.
\end{equation}
For the none-zero terms,  
\begin{align}
    \Dot{\boldsymbol{\xi}}_i &= 
    \Dot{\mathbf{R}}_{0i-1} {}_\mathcal{B}\mathbf{h}_i \times {}_\mathcal{A}\mathbf{r}_{iC} \\
    &+ \mathbf{R}_{0i-1} {}_\mathcal{B}\mathbf{h}_i \times _\mathcal{A}\Dot{\mathbf{r}}_{iC}
\end{align}
where $\Dot{\mathbf{r}}_{iC}$ can be computed through finite difference. The time derivative of the rotation matrix is computed as 
\begin{equation}
    \Dot{\mathbf{R}}_{0i-1} = (\mathbf{J}_{R,i-1}\Dot{\mathbf{q}})^\times \mathbf{R}_{0i-1}
\end{equation}
On the other hand, the time derivatives of the angular Jacobian can be computed similarly as 
\begin{equation}
    \Dot{\mathbf{J}}_{R,n} = \begin{bmatrix}
        \mathbf{0} & \mathbf{0} &
        \mathbf{0} &
        \Dot{\mathbf{R}}_{0i-1} {}_\mathcal{B}\mathbf{h}_i & \cdots
    \end{bmatrix}.
\end{equation}

In summary, Phys-SMPL computes the body mass and inertia information directly from SMPL specified by its shape parameters. The calculation of the physical terms in the Euler-Lagrange equations is also fully-differentiable, facilitating the seamless integration with physics and deep learning models. Additional background of the Euler-Lagrange equations can be found in \cite{RobotDynamics}.

\section{Selection of Body Contact Regions}
\label{supp:contactregion}

\begin{figure}[t]
    \centering
    \includegraphics[width=0.45\textwidth]{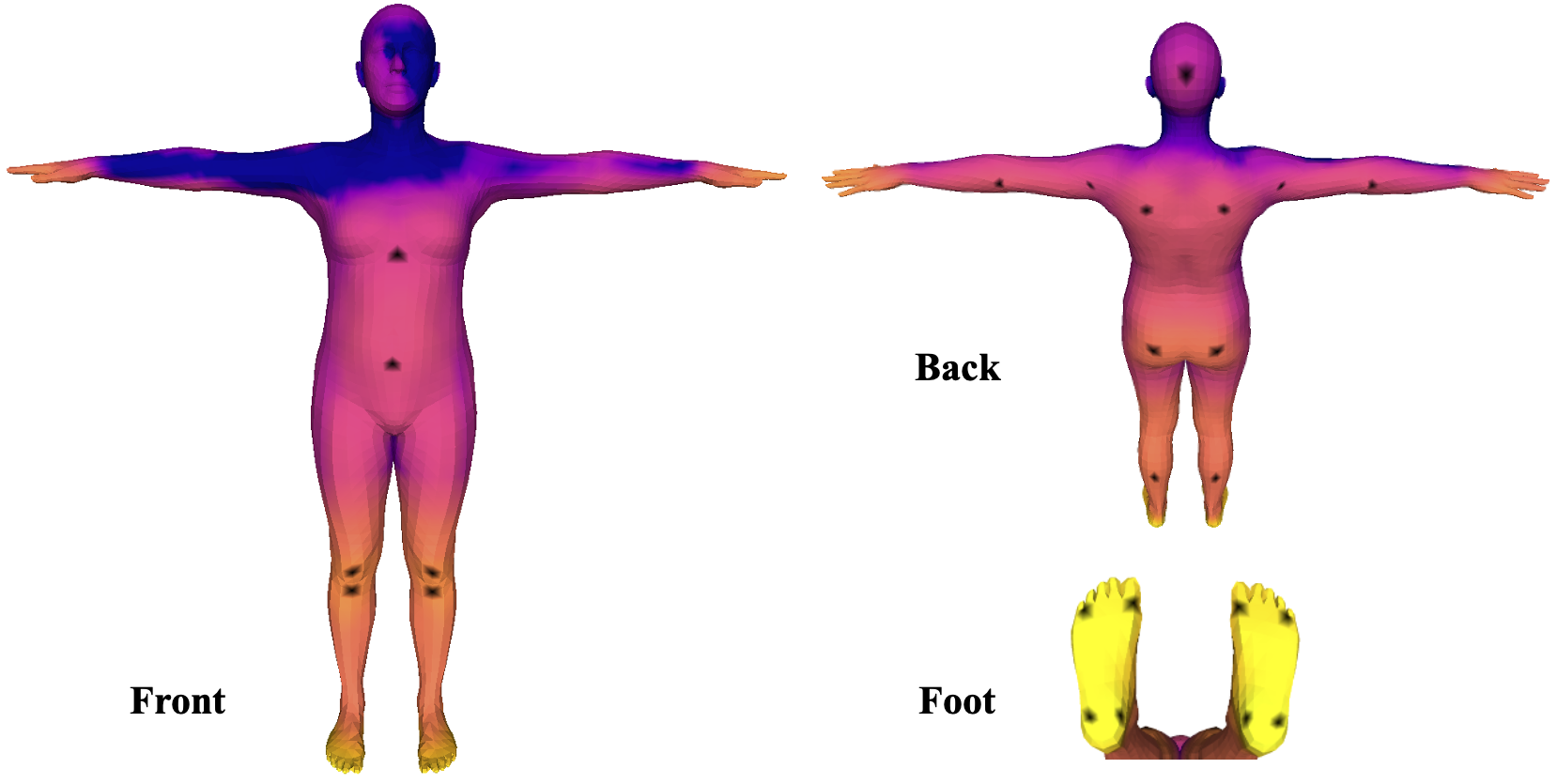}
    \caption{\textbf{Contact Map.} The lighter color (yellow) of a vertex indicates the higher frequency of contact with the ground. The modeled contact vertices are marked in black.}
    \label{fig:contactmap}
\end{figure}

We consider that the human body receives contact forces primarily from the ground. To effectively and efficiently model the contact behavior, we first investigate the body regions that frequently contact with the ground. We employ the motion sequence in AMASS and count the frequency of the vertices in contact with the ground. We visualize the results in Figure~\ref{fig:contactmap}. As expected, not all vertices have frequent contacts with the ground. On the other hand, modeling the contact for all vertices can result in significant computational overhead. Following existing approaches, we model a subset of vertices for each body part that frequently come into contact with the ground. The chosen vertices are highlighted by black colors in Figure~\ref{fig:contactmap}.

\section{Improvements over Different Kinematics-based 3D Body Reconstruction Models}
\label{supp:differentbackbones}

\begin{table}[t]
\tabcolsep=0.02in
    \begin{center}
        \begin{tabular}{ l cc  ccccc }
    \toprule
    \multirow{2}{*}{Method} & \multicolumn{2}{c}{\textit{Rec. Error}} & \multicolumn{5}{c}{\textit{Phys. Plausibility}} \\ \cmidrule{2-3} \cmidrule(lr){4-8} 
     & MJE & P-MJE & ACCL & VEL & FS & GP & BOS \\ \cmidrule{1-8}
    SPIN~\cite{kolotouros2019learning} & 66.2 & 40.8 & 18.1 & 8.2 & 8.8 & 12.5 & 23.7 \\
    +PoseBert~\cite{baradel2022posebert} & 64.3 & 41.8 & 5.3 & 4.1 & 9.2 & 15.7 & 26.1 \\
    +\textbf{{\ours}} (\textbf{Ours}) & \textbf{60.7} & \textbf{40.3} & \textbf{2.5} & \textbf{3.6} & \textbf{3.0} & \textbf{2.1} & \textbf{31.4} \\
    \cmidrule{1-8}
    IPMAN~\cite{tripathi20233d} & 63.1 & 41.0 & 17.2 & 7.8 & 8.6 & 11.9 & 28.6 \\
    +PoseBert~\cite{baradel2022posebert} & 62.2 & 41.5 & 5.3 & 4.2 & 9.1 & 11.9 & 29.8 \\
    +\textbf{{\ours}} (\textbf{Ours}) & \textbf{59.4} & \textbf{40.2} & \textbf{2.5} & \textbf{3.6} & \textbf{3.0} & \textbf{2.3} & \textbf{36.1}\\
    \bottomrule
    \end{tabular}
    \end{center}
    \caption{\textbf{Evaluation on Human3.6M using Different Kinematics-based 3D Reconstruction Models.} The results of other works are from their officially released models. BOS is measured in percentages, with larger values indicating better performance, while the other metrics prefer smaller values.}
\label{tab:backbones}
\end{table}

{\ours} can be seamlessly applied to different kinematics-based models to enhance their motion estimates. To demonstrate this, besides employing CLIFF (evaluation is discussed in the main manuscript), we here report the improvements over two recent image-based 3D human body reconstruction methods, SPIN and IPMAN. SPIN and IPMAN are both model-based 3D human body reconstruction models that directly predict the 3D body pose and shape parameters from input images. IPMAN can produce more stable 3D body pose estimates than SPIN due to the incorporation of an intuitive-physics loss during training. To further demonstrate the effectiveness of {\ours} and compare with IPMAN, we follow IPMAN and compute the Base of Support (BOS) metric to evaluate the physical plausibility in terms of pose stability. We present the results in Table~\ref{tab:backbones} and compare the improvements achieved by our approach with those produced by PoseBert. As demonstrated, when integrated with both SPIN and IPMAN, {\ours} effectively improves the reconstruction accuracy and substantially enhances the physical plausibility. For instance, when evaluating the reconstruction accuracy, adding {\ours} on top of SPIN shows a reduction in MJE from 66.2mm to 60.7mm, and on top of IPMAN, it experiences a decrease from 63.1mm to 59.4mm. In terms of physical plausibility, incorporating {\ours} with SPIN results in a reduction of 15.6 mm/frame$^2$ in ACCL, and with IPMAN, it exhibits a reduction of 14.7 mm/frame$^2$. The improvements achieved by {\ours} are also more pronounced than PoseBert. Furthermore, when assessing stability, SPIN initially exhibits poor performance with a lower BOS than IPMAN (23.7\% vs. 28.6\%). By integrating with {\ours}, SPIN can generate a higher BOS than IPMAN (31.4\% over 28.6\%). The improvements in stability are also evident when incorporating {\ours} with IPMAN. {\ours} significantly enhances the motion estimates by effectively leveraging the Transformer model and integrating physics principles and it applies to various kinematics-based 3D reconstruction models.

\section{Evaluation on Global Motion Recovery}
\label{supp:evalglobalmotion}

\begin{table}[t]
\tabcolsep=0.01in
    \begin{center}
        \begin{tabular}{ l c c c}
    \toprule
    \multirow{2}{*}{Method} & Training on & \multirow{2}{*}{G-MPJPE ($\downarrow$)}  & \multirow{2}{*}{G-MPVPE ($\downarrow$)}\\
    & Human3.6M & & \\\midrule 
    D\&D \cite{li2022d} & Yes & 525.3 & 533.9 \\ 
    \midrule
    \textbf{{\ours}} (\textbf{Ours}) & \textbf{No} & \textbf{335.7} & \textbf{343.5} \\
    \bottomrule
    \end{tabular}
    \end{center}
    \caption{\textbf{Evaluation on Global Motion Recovery.} The evaluation is on the test set of Human3.6M. The units of G-MPJPE and G-MPVPE are in mm.}
\label{tab:gmpjpe}
\end{table}

\begin{figure*}[t]
    \begin{center}
    \includegraphics[width=0.98\linewidth]{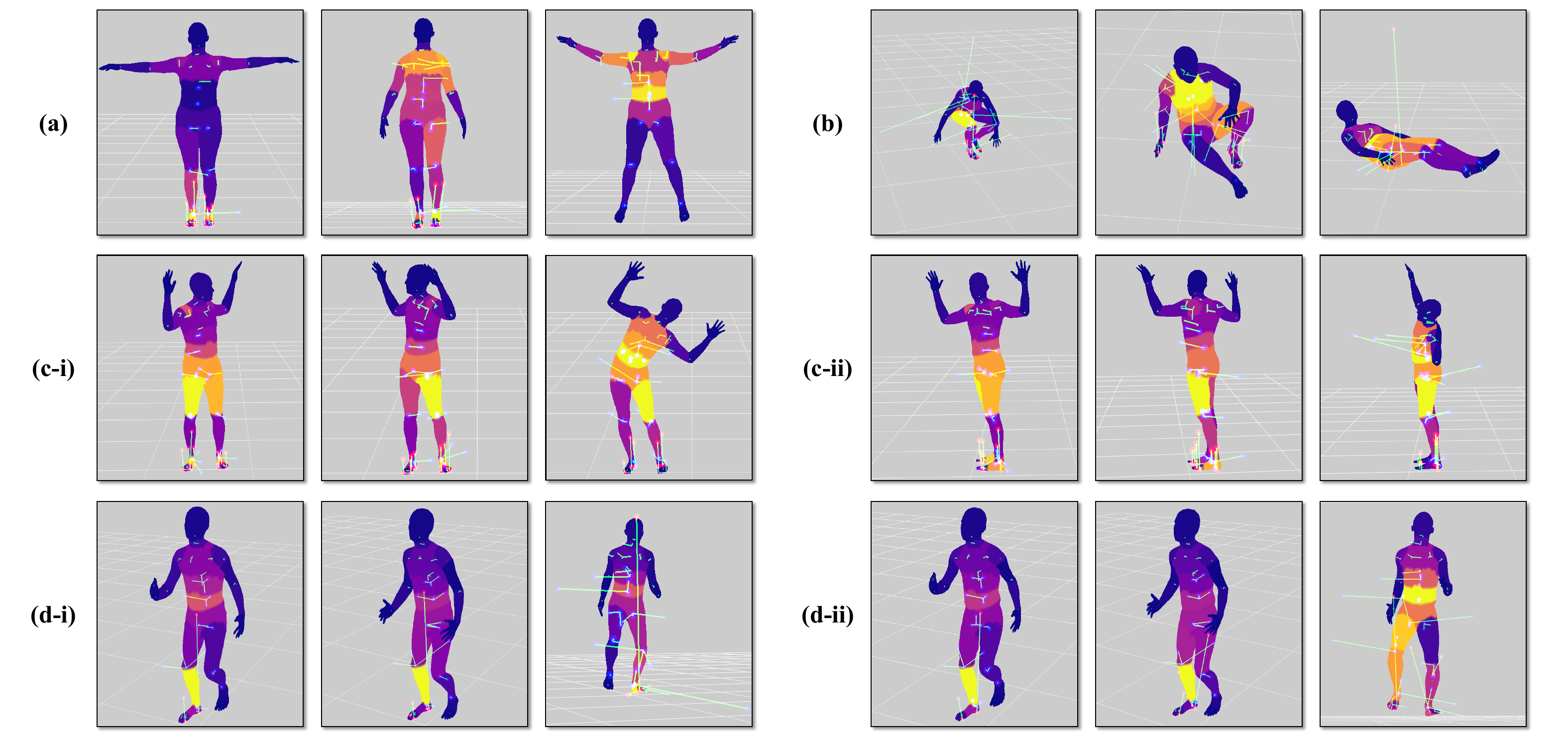}
    \end{center}
    \caption{\textbf{Visualization of the Inferred Motion Forces.} The motion sequences (a), (b), (c), and (d) are from AMASS~\cite{mahmood2019amass}. For sequence (c), (c-i) and (c-ii) are visualization of the same figure from different views. For sequence (d), (d-i) and (d-ii) are visualization of the forces generated with and without employing the continuous contact force model, respectively. In each image, the contact forces are visualized via a 3D vector on each contact point (green lines ended with red dots). The joint actuations are characterized by three vectors along the three Euler angles of a joint (green lines ended with blue dots). Meanwhile, the magnitudes of joint actuations are visualized by different colors at different body parts, with lighter colors indicating larger magnitudes.}
    \label{fig:forcelabel}
\end{figure*}

This section demonstrates the accurate global motion recovery achieved by our approach. We compute the Mean Per-Joint Position Errors and Mean Per-Vertex Position Errors in the world frame (G-MPJPE and G-MPVPE). Following the typical evaluation protocol~\cite{yuan2022glamr,li2022d}, we use a 10-second sliding window with the root translation aligned with the ground truth at the starting frame of each window. We report the results in Table~\ref{tab:gmpjpe} with comparison to the physics-based SOTA. As shown, our approach produces much lower errors  than the SOTA. For example, our approach achieves 335.7 mm of G-MPJPE, reducing D\&D's 525.3 by nearly 36\%. Moreover, it's worth noting that our method achieves better performance without utilizing any 3D data from Human3.6M during training.

\section{Quality of the Inferred Motion Forces}
\label{supp:pseudoforce}

The motion forces derived from the Euler-Lagrange equations using the physics-based body representation and the contact force model offer valuable insights into human dynamic behaviors. Utilizing them enables a more effective incorporation of the physics equations. We here discuss the quality of the inferred motion forces. In Figure~\ref{fig:forcelabel}, we present example forces generated from different motion sequences in AMASS. As illustrated, the inferred motion forces sensibly indicate the direction and magnitude of the underlying forces. For example, during normal standing (Figure~\ref{fig:forcelabel}-a, column 1), the contact forces are evenly distributed between both feet. When leaning to the left or right, the center of gravity shifts, and larger contact forces are displayed on the left or right foot accordingly (Figure~\ref{fig:forcelabel}-c, columns 1-2). Additionally, the contact forces applied to different body parts, such as those experienced on the feet and hips, are effectively modelled (Figure~\ref{fig:forcelabel}-b, columns 2-3). On the other hand, the inferred joint actuations clearly indicate the rotation direction and force magnitudes. For example, the spine joint actuation in the horizontal direction controls rotation along the horizontal axis. It changes direction when rotating the upper body from left to right (Figure~\ref{fig:forcelabel}-c, columns 1-2). Moreover, large forces are shown at the shoulder joints when extending the arms (Figure~\ref{fig:forcelabel}-a, columns 2-3), or at the spine joints when rotating the upper body (Figure~\ref{fig:forcelabel}-c, column 3).

To demonstrate the benefits of utilizing the continuous contact force model, we further compare the forces inferred with and without employing the contact force model. As illustrated in Figure~\ref{fig:forcelabel}-d, exploiting the contact force model results in a more stable estimation of the forces. Additionally, when not using the contact force model, a contact status must be determined beforehand in a heuristic manner. For example, a point is considered in contact if its distance to the ground is less than 3 cm, and its velocity is less than 1 m/s~\cite{yi2022physical}. In contrast, utilizing the contact force model eliminates the need for estimating the contact status and directly infers the contact forces based on a spring-mass model. Utilizing the contact force model can avoid the problems caused by incorrect estimations of the contact status (Figure~\ref{fig:forcelabel}-d-ii, column 3).

\begin{table*}[t]
\tabcolsep=0.01in
    \begin{center}
    \begin{tabular}{ l  ccccccccccccccc c}
    \toprule
    \multirow{2}{*}{Input} & \multirow[b]{2}{*}{\shortstack[c]{Baseball\\Pitch}} & \multirow[b]{2}{*}{\shortstack[c]{Clean\\\&Jerk}}& \multirow[b]{2}{*}{{\color{ForestGreen}\shortstack[c]{Pull\\Ups}}}& \multirow[b]{2}{*}{\shortstack[c]{Strum\\Guitar}}& \multirow[b]{2}{*}{\shortstack[c]{Baseball\\Swing}}& \multirow[b]{2}{*}{\shortstack[c]{Golf\\Swing}}& \multirow[b]{2}{*}{\shortstack[c]{Push\\Ups}}& \multirow[b]{2}{*}{\shortstack[c]{Tennis\\Forehand}}& \multirow[b]{2}{*}{{\color{ForestGreen}\shortstack[c]{Bench\\Press}}}& \multirow[b]{2}{*}{\shortstack[c]{Jumping\\Jacks}}& \multirow[b]{2}{*}{\shortstack[c]{Sit\\Ups}}& \multirow[b]{2}{*}{\shortstack[c]{Tennis\\Serve}}& \multirow{2}{*}{Bowling} & \multirow[b]{2}{*}{\shortstack[c]{Jump\\Rope}}& \multirow{2}{*}{{\color{ForestGreen}Squats}} & \multirow{2}{*}{All}\\ 
    & &&&&&&&&&&&&&& \\\midrule 
    $\mathbf{J}_{phys}$ & 100.0 & 95.6 & 94.9 & 100.0 & 98.3 & 100.0 & 99.0 & 93.5 & 91.4 & 98.2 & 98.0 & 97.1 & 96.4 & 97.6 & 94.9 & 96.8 \\
    $\mathbf{F}$ & 100.0 & 93.3 & 97.0 & 97.8 & 96.6 &97.4& 98.1& 81.8& 94.3& 96.4& 90.0 & 92.9 & 88.1 & 97.6 & 95.9 & 94.4 \\
    $\mathbf{J}_{phys}$+$\mathbf{F}$ & 100.0 & 100.0 & 97.0 & 100.0 & 98.3 & 100.0 & 98.1 & 96.1 & 97.1 & 98.2 & 98.0 & 97.1 & 97.6 & 97.6 & 96.9 & 98.0 \\
    \bottomrule
    \end{tabular}
    \end{center}
    \caption{\textbf{Action-wise Evaluation on PennAction Utilizing Different Model Inputs.} The numbers represent recognition accuracy in percentages. Actions that show higher accuracy when utilizing forces ($\textbf{F}$) compared to utilizing the physics-based estimation of 3D body joint positions ($\textbf{J}_{phys}$) are marked in green. The term ``All" denotes the average accuracy over all actions.}
\label{tab:actionwise}
\end{table*}

\section{Action-wise Recognition Performance}
\label{supp:actionrecognition}

In the main manuscript, we demonstrate that combining motion and force estimates leads to the best model performance. In this section, we discuss the action-wise evaluation results, providing further insights into the benefits of utilizing forces for understanding human behaviours. 

\noindent\textbf{Action-wise Evaluation.} The action-wise evaluation results are summarized in Table~\ref{tab:actionwise}. As shown, only using forces as model input produces a lower average recognition accuracy compared to utilizing the physics-based motion estimates. Nonetheless, utilizing forces yields higher accuracy for certain actions, such as ``Pull Ups", ``Bench Press", and ``Squats". These actions are distinctive particularly in their underlying motion forces. Specifically, "Pull Ups" and ``Bench Press" involve similar body movements, such as raising the arms with bending legs. However, their underlying motion forces are significantly different. ``Pull Ups" involves body lifting, while ``Bench Press" involves body lying on a bench. For these actions, the estimated forces provide additional insights towards the human dynamic behaviours, and when combined with the 3D position data, they can significantly improve the final recognition accuracy. For example, the recognition accuracy of ``Bench Press" increases from 91.4\% to 97.1\% by further adding the forces as model input.

We here present the implementation details for reproducibility. \textbf{Implementation Details.} For our experiments on PennAction, we follow the established protocol that use the official split to divide the training and test sets. The skeleton graph is defined in the SMPL joint format. For the motion estimate input, they are 3D body joint positions. For the force input, they are estimated force values, where the contact forces are transformed into the generalized coordinates to align with the defined skeleton graph. When combining the motion and force estimates, we employ a decision-level fusion. The models are trained by minimizing the cross-entropy loss. We utilize the Adam optimizer with a weight decay of $10^{-4}$. We train the models for 200 epochs with an initial learning rate of $10^{-4}$ and decreasing to its 0.8 after every 40 epochs. The batch size is 128.